\documentclass[a4paper,journal,twoside,onecolumn]{IEEEtran}

\pdfoutput=1

\usepackage{amsmath,amsfonts}
\usepackage{algorithmic}
\usepackage{array}
\usepackage[caption=false,font=normalsize,labelfont=sf,textfont=sf]{subfig}
\usepackage{textcomp}
\usepackage{stfloats}
\usepackage{url}
\usepackage{verbatim}
\usepackage{graphicx}
\usepackage{balance}

\usepackage{scalerel}
\usepackage{tikz}
\usetikzlibrary{svg.path}

\definecolor{orcidlogocol}{HTML}{A6CE39}
\tikzset{
	orcidlogo/.pic={
		\fill[orcidlogocol] svg{M256,128c0,70.7-57.3,128-128,128C57.3,256,0,198.7,0,128C0,57.3,57.3,0,128,0C198.7,0,256,57.3,256,128z};
		\fill[white] svg{M86.3,186.2H70.9V79.1h15.4v48.4V186.2z}
		svg{M108.9,79.1h41.6c39.6,0,57,28.3,57,53.6c0,27.5-21.5,53.6-56.8,53.6h-41.8V79.1z M124.3,172.4h24.5c34.9,0,42.9-26.5,42.9-39.7c0-21.5-13.7-39.7-43.7-39.7h-23.7V172.4z}
		svg{M88.7,56.8c0,5.5-4.5,10.1-10.1,10.1c-5.6,0-10.1-4.6-10.1-10.1c0-5.6,4.5-10.1,10.1-10.1C84.2,46.7,88.7,51.3,88.7,56.8z};
	}
}

\newcommand\orcidicon[1]{\href{https://orcid.org/#1}{\mbox{\scalerel*{
				\begin{tikzpicture}[yscale=-1,transform shape]
					\pic{orcidlogo};
				\end{tikzpicture}
			}{|}}}}

\usepackage{hyperref} 

\newcommand{\orcidSara}{\textsuperscript{\orcidicon{0000-0002-6433-6510}}} 
\newcommand{\orcidMario}{\textsuperscript{\orcidicon{0000-0003-0353-3905}}} 
\newcommand{\orcidNacho}{\textsuperscript{\orcidicon{0000-0001-9704-0022}}} 
\newcommand{\orcidRafa}{\textsuperscript{\orcidicon{0000-0001-8486-1032}}} 

\begin{document}
\title{Reinforcement Learning Approach to Optimizing Profilometric Sensor Trajectories for Surface Inspection}


\author{Sara Roos-Hoefgeest\orcidSara, 
	    Mario Roos-Hoefgeest\orcidMario,
	    Ignacio Alvarez\orcidNacho,
	    and Rafael C. González\orcidRafa \IEEEmembership{Member,~IEEE}
	\thanks{Sara Roos-Hoefgeest, Ignacio Alvarez and Rafael C. González are with the Department of Electrical, Electronic, Communications and Systems Engineering, Oviedo University, Gijón, Spain}
	\thanks{Mario Roos-Hoefgeest is with CIN Systems, Gijón, Spain}
}

\markboth{AUGUST 2024}{Roos-Hoefgeest et al.: RL Approach to Optimizing Trajectories Surface Inspection}

\maketitle

\begin{abstract}

High-precision surface defect detection in manufacturing is essential for ensuring quality control. Laser triangulation profilometric sensors are key to this process, providing detailed and accurate surface measurements over a line. To achieve a complete and precise surface scan, accurate relative motion between the sensor and the workpiece is required, typically facilitated by robotic systems. It is crucial to control the sensor's position to maintain optimal distance and orientation relative to the surface, ensuring uniform profile distribution throughout the scanning process. Reinforcement Learning (RL) offers promising solutions for robotic inspection and manufacturing tasks. This paper presents a novel approach using RL to optimize inspection trajectories for profilometric sensors. Building upon the Boustrophedon scanning method, our technique dynamically adjusts the sensor's position and tilt to maintain optimal orientation and distance from the surface,  while also ensuring a consistent profile distance for uniform and high-quality scanning. Utilizing a simulated environment based on the CAD model of the part, we replicate real-world scanning conditions, including sensor noise and surface irregularities. This simulation-based approach enables offline trajectory planning based on CAD models.
Key contributions include the modeling of the state space, action space, and reward function, specifically designed for inspection applications using profilometric sensors. We use Proximal Policy Optimization (PPO) algorithm to efficiently train the RL agent, demonstrating its capability to optimize inspection trajectories with profilometric sensors. 
To validate our approach, we conducted several experiments where a model trained on a specific training piece was tested on various parts in simulation. Also, we conducted a real-world experiment by executing the optimized trajectory, generated offline from a CAD model, to inspect a part using a UR3e robotic arm model.

\end{abstract}

\def\abstractname{Note to Practitioners}
\begin{abstract}
	This paper addresses the problem of integrating robot manipulators and laser profilometric sensors in surface inspection tasks. Despite the relevance of both technologies in the manufacturing industry, little research has been done to integrate both technologies. In this work, we present a Reinforcement Learning-based trajectory generation algorithm that generates feasible sensor trajectories that optimizes the sensor’s position and orientation at each point along the scanning path. The input to the system is a CAD model of the object to be inspected. Therefore the system adapts dynamically to the geometrical characteristics of the objects. Experiments were conducted using simulations and real hardware. Objects with different geometric shapes and complexities were used to validate the approach, proving the effectiveness of this approach.
\end{abstract}

	
\section{Introduction}

\IEEEPARstart{S}{urface} inspection is a critical aspect of quality control in many industries, ensuring that manufactured components meet strict standards and function reliably. Accurate detection and characterization of surface defects are essential to maintaining product integrity and quality. 

Many industries still rely on manual inspection processes performed by human operators. However, manual inspection has already stopped being practical when it comes to the development of industrial demands for accuracy and efficiency. Manual inspection systems are not prone to detecting micron-scale defects. Therefore, advance sensor technologies are needed to serve for the need of small imperfections, which would be impossible to detect by human vision. A famous case study is car body pieces \cite{Zhou2019}. As the body components have a dimensional tolerance of several tenths of millimeter, the defects, whether it is functional stretching, or just a purely aesthetic protrusion, or imbalance, are sometimes one hundredth of this size. 

Furthermore, to meet the stringent requirements of modern industrial inspection, advanced sensor technologies have emerged as indispensable tools. Among them, laser triangulation is a widely adopted technique due to its superior precision and efficiency \cite{7801541}, \cite{laserTriang}. This method involves projecting a laser line onto the surface of an object and capturing the reflected light using a camera or sensor. Through the analysis of the distortion of the projected line, detailed information about the surface topography can be obtained with high accuracy.

To achieve an integral scan of the entire surface of the part to be inspected, relative motion between the part and the sensor is required. Robotic systems, including robotic arms, \cite{Khan2021}, \cite{OAKI20239354}, unmanned aerial vehicles (UAVs) \cite{drones5040106}, drones or unmanned ground vehicles (UGVs) \cite{inproceedings}, \cite{s20216384}, or autonomous underwater vehicles (AUVs) \cite{submarinos}, have been increasingly integrated into inspection procedures in different applications to address this requirement. These systems facilitate precise and controlled travel between the inspected part and the sensor, enabling complete surface coverage and efficient inspection processes. 

Effective and accurate inspection requires meticulous planning of the sensor paths over the part surface. While manual planning is sufficient for simpler scenarios, more complex geometries or stringent accuracy standards require the implementation of automated methods. The generation of inspection paths for robotic systems represents a significant challenge, requiring predefined paths that take into account surface geometry, defect characteristics and inspection requirements.

Although several studies related to automated inspection path planning can be found in the literature, highlighting the use of robotic arms, there is a significant gap in research specifically addressing the integration of robotics and profilometric sensors for surface inspection tasks.

Chen et al. highlight this gap in their study\cite{Chen2023}, where they propose a novel approach for automatically detecting surface defects on freeform 3D objects using a 6-degree-of-freedom manipulator equipped with a line scanner and a depth sensor. Their method involves defining local trajectories for precise defect inspection and optimizing global time for efficient scanning. 

Li et al. propose a method for planning scanning trajectories in automated surface inspection \cite{Li2019}. Their approach is based on a trajectory planning algorithm using a triangular mesh model. They divide the workpiece surface area into regions, determine scanning orientations and points in each region, and then generate scanning trajectories using the minimum enclosing rectangle. This method involves developing a section division algorithm to determine the scanning orientation, followed by generating trajectories that comply with system constraints.

Recently, a new trend has emerged for trajectory generation in robotics using Reinforcement Learning (RL) methods. While several papers have explored the potential of RL in various applications, few have focused specifically on its application in inspection tasks. This lack of research highlights the need for further exploration and research in this area.

For example, Lobbezoo et al. compile in \cite{Lobbezoo2021} different strategies present in the literature that use RL algorithms for pick and place applications. On the other hand, Elguea-Aguinaco et al. provides in \cite{InigoReview} a comprehensive review of the current research on the use of RL in tasks involving intensive physical interactions. These tasks refer to activities where object manipulation involves direct and meaningful contact between the robot and its environment. This study covers research related to a variety of areas, including rigid object manipulation tasks (e.g., assembly, disassembly, or polishing and grinding) and deformable object manipulation tasks (e.g., rope or garment and fabric folding, tensioning and cutting, or object manipulation). The approaches and methodologies employed in each of these areas are compiled and analyzed.

Han et al. present in \cite{Han2023} a comprehensive investigation of different applications of Deep RL in robotic manipulators, highlighting the main problems faced by RL in robotic applications. One of the most important problems is that the models used often do not perfectly replicate the real system. For example, in machine vision-guided robots, the simulation of RGB images can differ significantly from the actual images captured by cameras, a problem known as Sim-to-Real. This is because the simplified models do not fully capture the system dynamics. These discrepancies make it difficult to transfer simulation-trained models to real environments.

To address this issue, we use a realistic simulator presented in our previous work \cite{simu_roos}, which allows us to accurately represent the measurements obtained by the profilometric laser triangulation sensor. 

Another problem they highlight is that trajectory generation in robotics is an inherently multidimensional problem, which further complicates the learning and optimization process. Also Ji et al. emphasize in \cite{8809005} this problem, highlighting that in the field of robotics, most work using RL focuses on the field of mobile robot navigation due to its simple and well-developed theory \cite{9409758}.

Surface inspection using profilometric sensors is typically performed in a straight line. If the workpiece is too large to be covered in a single scan, parallel passes are used, generally following Boustrophedon-like trajectories \cite{GALCERAN20131258}. In our approach, we will start with these linear or Boustrophedon trajectories and enhance them using reinforcement learning techniques to optimize sensor movements and ensure comprehensive surface coverage. During each pass, the sensor advances in a predetermined direction while it can adjust its height and pitch angle over the piece, keeping the other orientations constant. Our approach effectively tackles the multidimensional challenge in surface inspection by concentrating on three critical parameters: the sensor's position along the scanning direction, its height adjustment, and pitch orientation. This focus simplifies the problem significantly, eliminating the need to individually manage each robot axis and streamlining sensor control and trajectory planning during inspections.

Reinforcement learning (RL) techniques offer effective solutions for problems with limited action spaces, making them well-suited for optimizing surface inspection tasks. RL's ability to learn from interactions and improve control policies based on rewards makes it a promising tool for addressing challenges in this field. Despite advancements in RL algorithms, its application in inspection tasks remains relatively underexplored compared to other robotics applications. This gap highlights the necessity for further exploration and research in this area.

In the realm of inspection applications, Xiangshuai Zeng's research presents PIRATE (Pipe Inspection Robot for Autonomous Exploration) \cite{Zeng2019ReinforcementLB}, a robot designed for internally inspecting pipes using reinforcement learning. Equipped with multiple bending joints and wheels, PIRATE offers adaptive and efficient mobility. By employing algorithms like PPO and deep neural networks, the robot learns to navigate sharp corners, adjusting its behavior in real-time and adapting to various pipe configurations. The system defines actions, rewards, and observations, with actions including wheel movements and adjustments in bending joints, and observations coming from a 2D laser scanner.

Another work focused on inspection applications is presented by Jing et al. in \cite{Jing2018}, centering on the automatic generation of robotic trajectories for surface inspection in production lines. They use techniques such as Monte Carlo algorithms and greedy search for Coverage Path Planning (CPP), enabling the automatic generation of inspection trajectories adapted to objects of different sizes and geometries. Using a 3D structured light scanner, they generate a 3D point cloud representing the scanned object's geometry. The main objective is to minimize total cycle time, combining View Planning Problem (VPP) and trajectory planning to minimize the total sum of inspection and displacement costs after meeting surface coverage requirements. The trajectory generation process includes random viewpoint selection, robot movement calculation, collision-free path planning, evaluation of visibility for each viewpoint and covered surface portion, and application of an RL-based planning algorithm to select inspection actions until completion. Actions involve selecting a viewpoint and moving the robot to place the 3D scanner at the selected viewpoint. The proposed RL algorithm automatically generates the online inspection policy, taking as input the robot model, target object model, and sensor specifications.

Aligned with prior research, Christian Landgraf, Bernd Meese et al. introduce in \cite{Landgraf2021} an approach for automatic viewpoint planning in surface inspection. They propose a strategy to find optimal sets of viewpoints for 3D inspection of specific workpieces. Their goal is to automate this process for any industrial robotic arm available in ROS and any 3D sensor specification; in their application they employ a stereo camera. Using advanced reinforcement learning algorithms such as Q-learning, Proximal Policy Optimization (PPO) and Deep Q-Networks (DQN), each action involves selecting the viewpoint and planning and executing the robot's movement towards this pose. Upon reaching its goal, the sensor generates a 3D point cloud at this specific pose, with the state of the environment constructed using 3D measurement observations and the robot's current pose.

For case studies like the one in this article, which uses laser triangulation profilometric sensors for measurements along a line, traditional trajectory planning approaches, such as the mentioned View Planning Problem (VPP), are not suitable. The VPP is intended for finding optimal poses for capturing images or making measurements of objects or surfaces using 3D vision sensors that cover a wide area. This highlights the need for a trajectory planning approach adapted to the specific characteristics of profilometric sensors, focusing on optimizing surface exploration rather than capturing a complete three-dimensional view.

To date, no research has been found regarding the generation of inspection trajectories using Reinforcement Learning and profilometric sensors, highlighting a significant gap in the literature and a promising area for robotics and automation research. This study addresses this gap by presenting an RL-based approach designed for generating inspection trajectories using profilometric sensors. Key contributions involve the modeling of the state space, action space, and reward function. We use the PPO (Proximal Policy Optimization) algorithm to efficiently train the RL agent, demonstrating its ability to optimize inspection trajectories with profilometric sensors.

PPO is an algorithm proposed by OpenAI in \cite{SchulmanWDRK17}. The authors highlight its ability to balance three key aspects in reinforcement learning: ease of implementation, sampling efficiency, and simplicity in hyperparameter tuning. PPO not only offers competitive performance comparable to or better than more advanced reinforcement learning algorithms, but also stands out for its significantly simpler implementation and tuning.

The effectiveness of PPO has been demonstrated across a wide range of applications, including robot control such as the PIRATE robot \cite{Zeng2019ReinforcementLB} or the development of view planning systems for inspection, as presented in the work of Landgraf, Meese et al. \cite{Landgraf2021}. Additionally, it has been successfully applied in pick-and-place tasks, such as training 7-degree-of-freedom robotic arms for precise object manipulation, as described in \cite{9282951}, as well as in other pick-and-place applications, as discussed in \cite{robotics12010012}.

Furthermore, comparisons between PPO and other reinforcement learning algorithms, such as SAC and TD3, reveal interesting patterns in terms of training efficiency, performance, and convergence. For example, in \cite{Mock2023}, it was found that PPO tends to perform better in smaller state spaces, while SAC shows advantages in larger state spaces. On the other hand, in \cite{robotics12010012}, PPO and SAC were compared, where SAC demonstrated greater efficiency in sampling, but PPO exhibited greater insensitivity to hyperparameters and more stable convergence in complex problems. These findings support the choice of PPO as the main algorithm for the proposed research.


\section{Materials and Methods}

In this section, we present the proposed approach for generating inspection trajectories using profilometric sensors and Reinforcement Learning techniques. The goal is to optimize each pass of the Boustrophedon scanning method, seeking an optimal orientation and distance of the sensor relative to the part at each position. This involves dynamically adjusting the position and tilt (pitch) of the sensor to maintain a constant and appropriate pose between the sensor and the surface at all times. The other two sensor orientations will be fixed beforehand, allowing for precise and uniform data capture. Additionally, we aim to optimize the spacing between profiles. This approach ensures complete and detailed coverage of the part's surface, thereby maximizing the quality and accuracy of the captured data.

To train the RL algorithms, we use a simulated environment that replicates the conditions of the real system described developed in our previous work \cite{simu_roos}. This simulator emulates the measurements of a laser triangulation profilometric sensor, including sensor noise and speckle noise generated by the object's surface. Thus, a realistic and controlled training environment is obtained.

The state space is constructed using the position and orientation of the robot's end-effector. This allows for generalization of the approach and facilitates the transfer of the method to different robotic configurations. Additionally, the state also includes other parameters, such as the mean profile distance, the direction angle, and the advance between consecutive scans.

The action space is defined by relative increments in the sensor's position and tilt angle, allowing for precise adjustments and smooth movements of the sensor. The reward function consists of three key components: the distance between the sensor and the surface, the alignment of the sensor with the surface normal, and the spacing between consecutive scans. This comprehensive reward function encourages optimal behaviors in terms of distance, orientation, and sensor advancement.

Next, each component of the proposed method is described in detail, providing a thorough understanding of its design and implementation.

\subsection{Reinforcement Learning (RL): Basic Concepts}

Reinforcement Learning (RL) is a branch of machine learning inspired by behavioral psychology, which focuses on how agents make decisions within an environment to maximize some measure of accumulated reward over time \cite{Sutton1998}. 

\begin{figure}
    \centering
    \includegraphics[width=0.8\columnwidth]{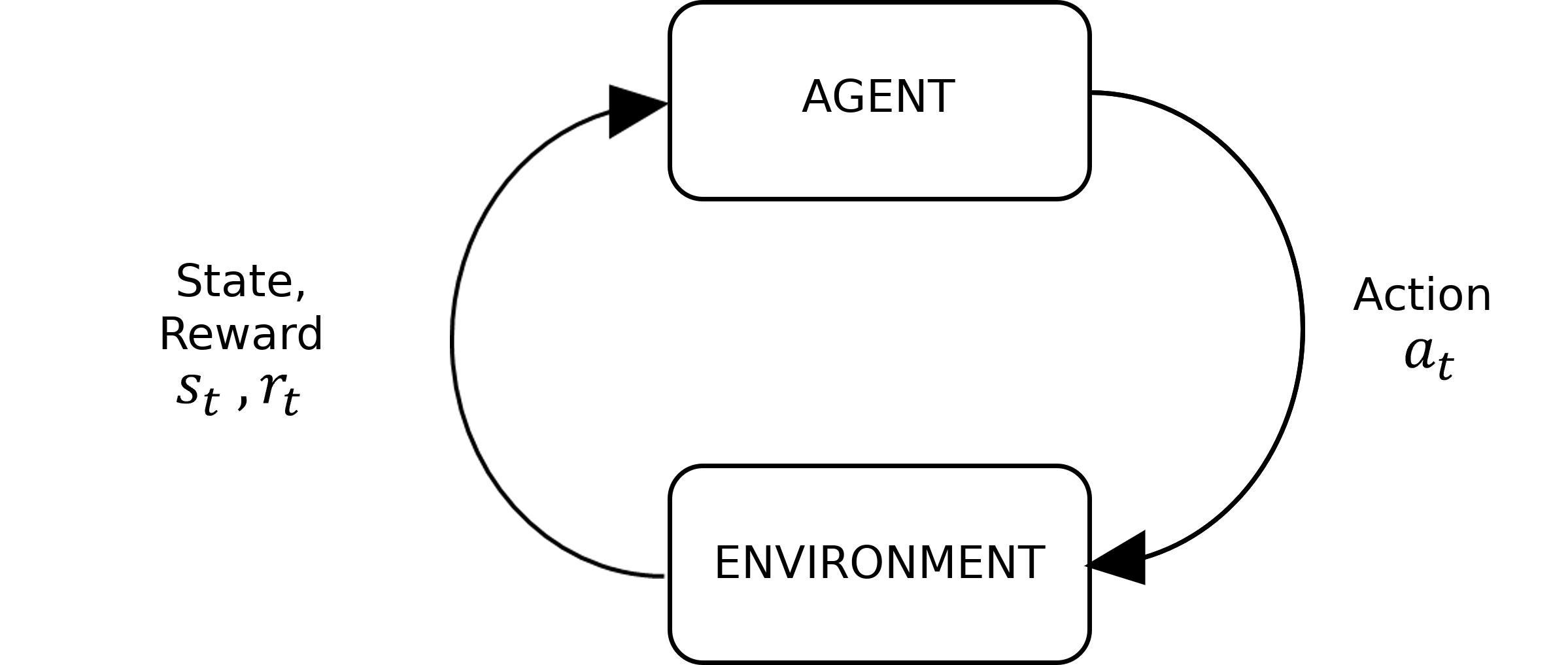}
    \caption{Agent-Environment Interaction Cycle}
    \label{fig:RLciclo}
\end{figure}

In RL, the agent dynamically interacts with the environment, observing its current state and selecting actions in response. These actions affect the environment's state and generate a reward signal that guides the agent's behavior. The primary goal of the agent is to maximize the accumulation of these rewards over time, thereby optimizing its performance in the environment. Figure \ref{fig:RLciclo} illustrates the basic interaction cycle between an agent and the environment.

The agent refers to the machine learning algorithm that interacts with the environment. The environment is the adaptive problem space with attributes such as variables, boundary values, rules, and valid actions. Each action represents a step taken by the RL agent to navigate through the environment. The set of all valid actions in a given environment is referred to as the action space $\mathcal{A}$, defined mathematically as $\mathcal{A} = \{a_1, a_2, ..., a_n\}$, where $a_i$ represents a specific action in the set, and $n$ is the total number of actions. A state represents the environment at a given point in time. The set of all possible states in an environment is $\mathcal{S} = \{s_1, s_2, ..., s_m\}$. The reward is the positive, negative, or neutral value the agent receives as a consequence of an action, assessing its quality. The reward at each time step $r_t$ depends on each state-action pair $r_t = r(s_t, a_t)$. The accumulated reward is the sum of all rewards obtained over time.

In most cases, the problem to be solved is formally modeled as a Markov Decision Process (MDP). An MDP can be defined as a tuple of 5 elements $(\mathcal{S}, \mathcal{A}, r, P, \rho_0)$, representing respectively the set of all valid states, the set of all valid actions, the reward function, the state-action transition probability function, and the initial state distribution. 

Another key parameter is the policy ($\pi$), which is the strategy guiding the agent's decision-making within the environment. The policy can be deterministic, where for each state in the environment, the agent selects a specific action predictably or stochastic, meaning that the agent chooses actions based on probabilities, introducing uncertainty and allowing for exploration.

The goal of any reinforcement learning algorithm is to select a policy that maximizes the expected return when the agent acts according to it. The expected return $J(\pi)$ is represented by equation \ref{eq:RLobj}. The expected reward $\mathbb{E}_{\pi}[r(s_t, a_t)]$ is the average of the rewards the agent expects to receive by following the policy $\pi$ in each state $s$. The objective is to adjust the policy parameters to maximize this reward, using optimization methods such as policy gradient to continuously improve the policy and the agent's performance in the specified task.

\begin{equation}
    J(\pi) = \mathbb{E}_{\pi}[r(s_t,a_t)]
    \label{eq:RLobj}
\end{equation}

Therefore, the final optimization problem in an RL algorithm can be expressed as equation \ref{eq:optRL}, where $\pi^*$ is the optimal policy:

\begin{equation}
    \pi^* = \arg \max_\pi J(\pi)
    \label{eq:optRL}
\end{equation}

\subsection{Scanning Characteristics for Surface Inspection with Profilometric Sensors}

During the scanning process using laser triangulation profilometric sensors, the quality of the measured data is directly affected by various parameters associated with the relative position between the sensor and the inspected piece, as detailed in \cite{Li2019}. These parameters are critical to ensuring precise and thorough surface inspection. Therefore, it is essential to carefully consider these factors during the planning of scanning trajectories in order to achieve effective results in surface inspection. 

    \begin{figure}
    \centering
    \subfloat[]{
        \includegraphics[width=0.25\columnwidth]{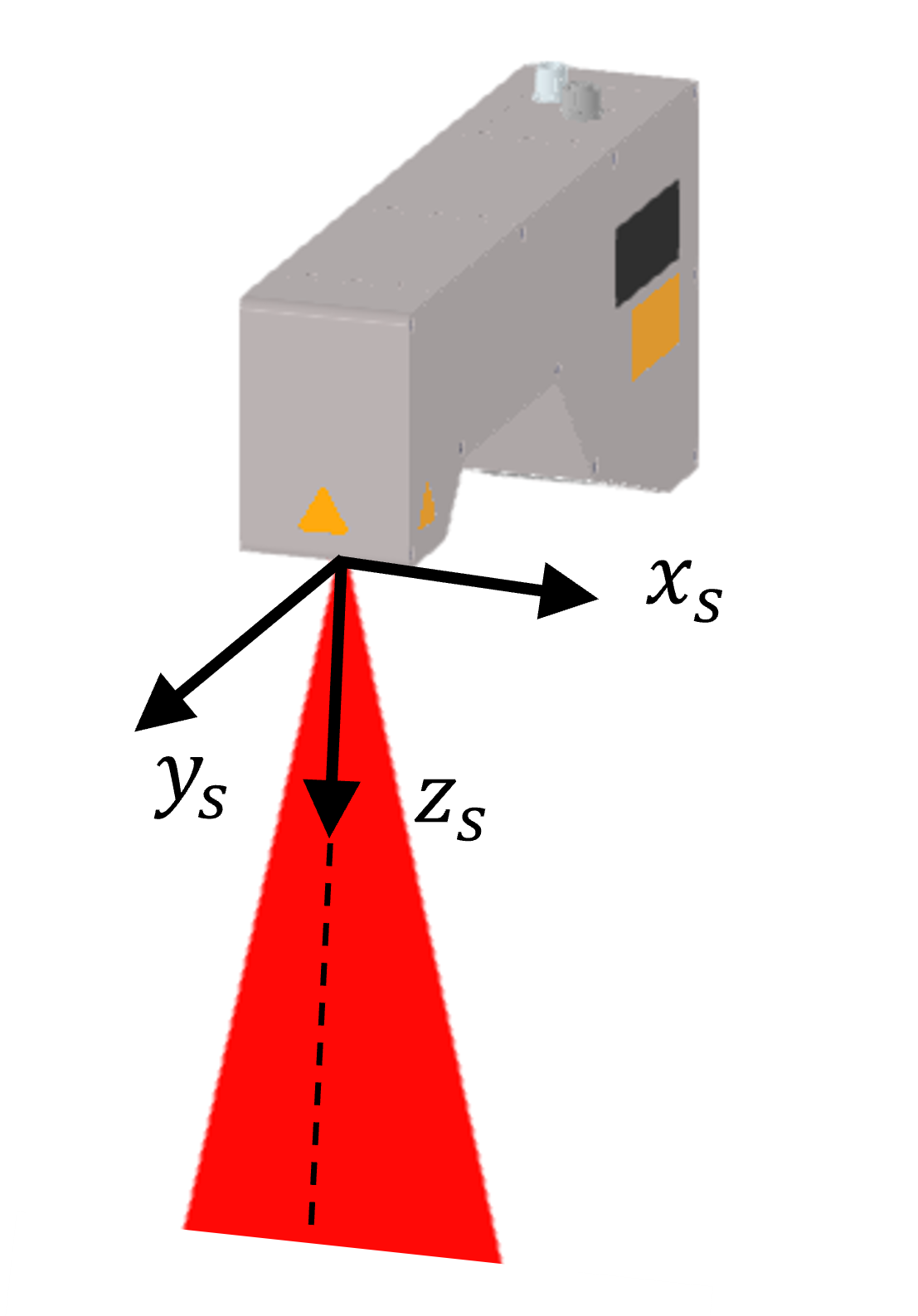}}
    \subfloat[]{
             \includegraphics[width=0.3\columnwidth]{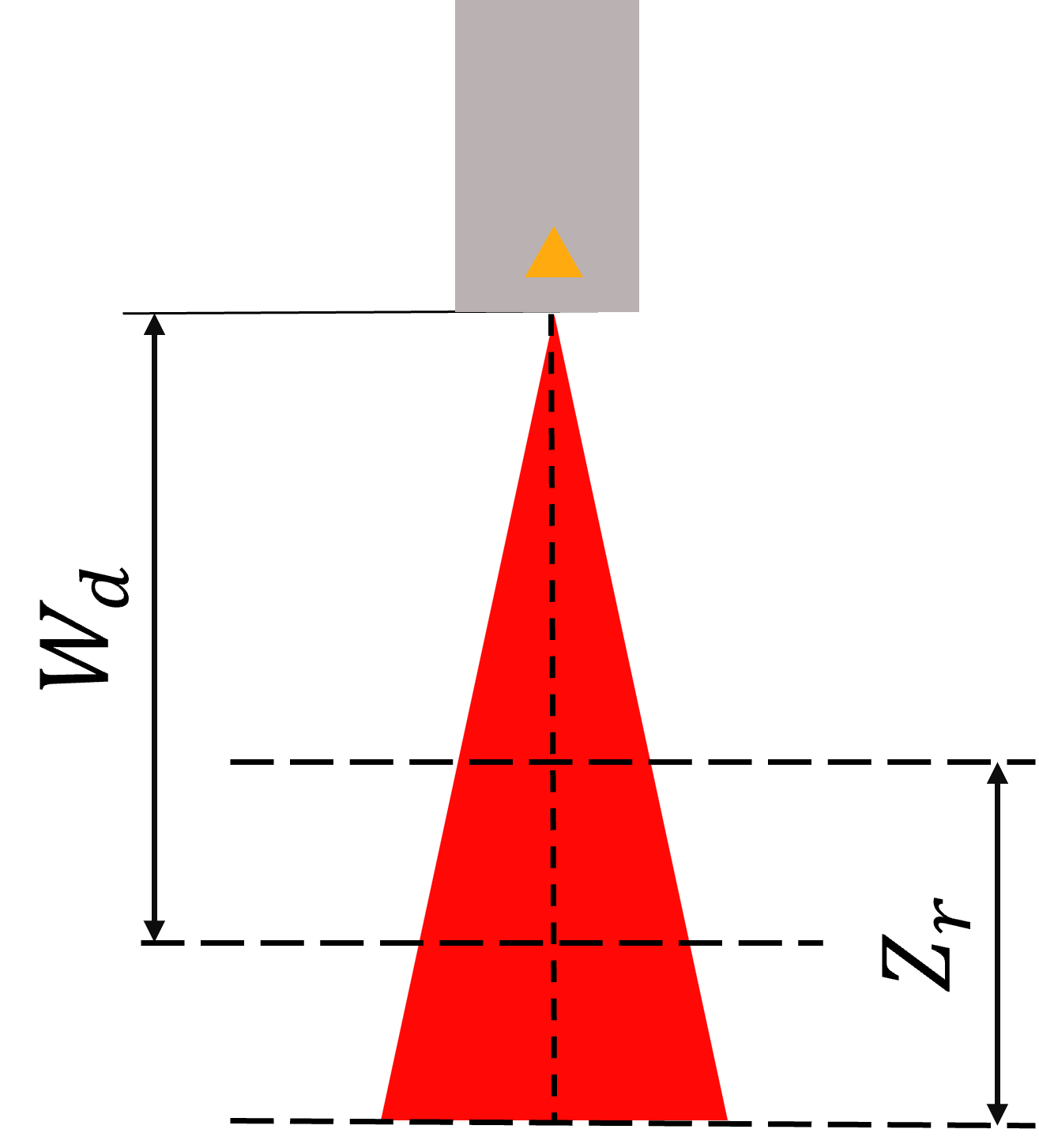}}
    \\
    \subfloat[]{
       \includegraphics[width=0.33\columnwidth]{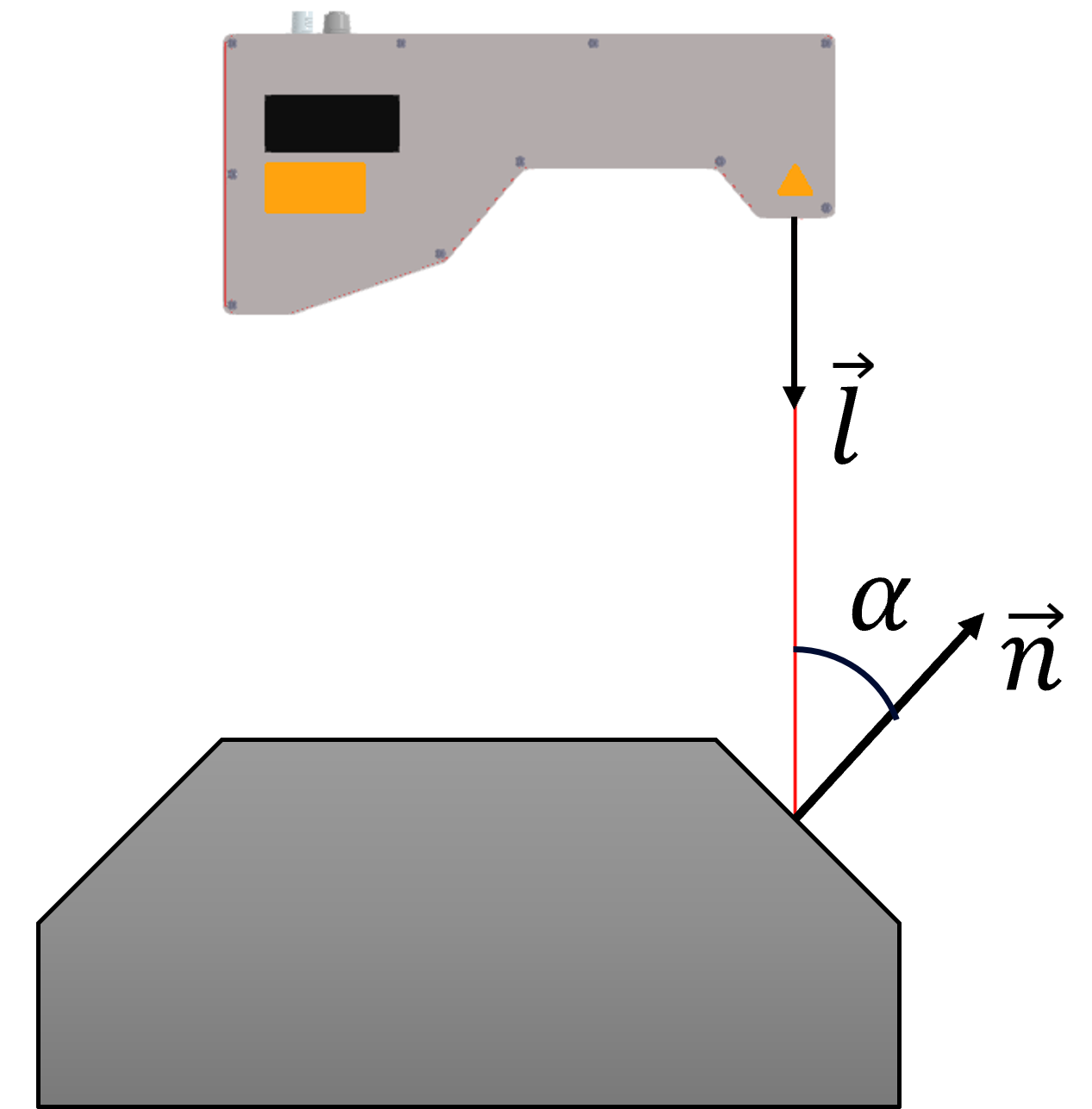}}
    \subfloat[]{
       \includegraphics[width=0.38\columnwidth]{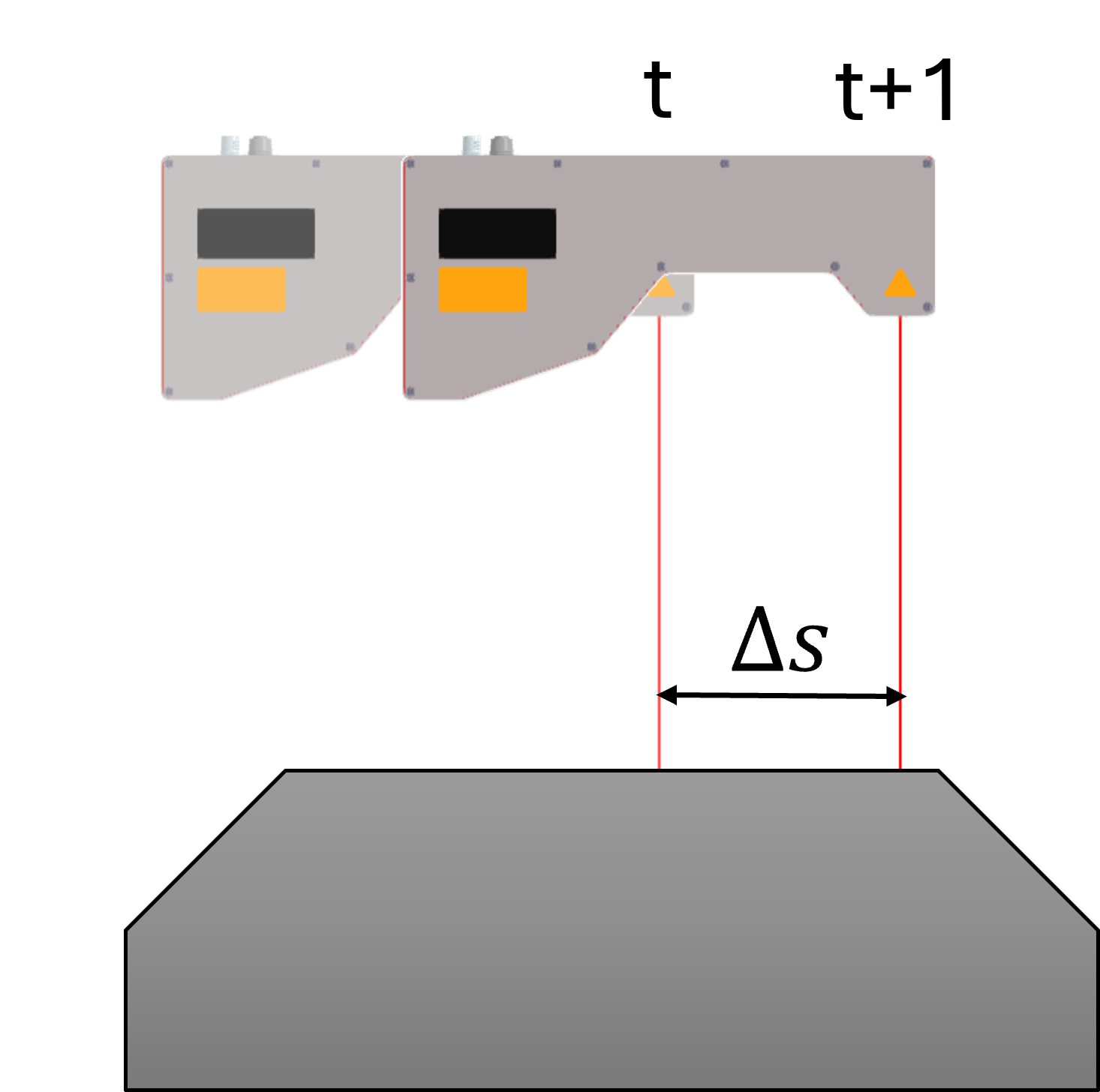}}
    \caption{Representation of the profilometric sensor and its main parameters. (a) 3D representation of the sensor with its coordinate system.
    (b) Frontal view of the sensor. The optimal working distance $W_d$ and depth of field $Z_r$ are depicted. (c) Representation of the direction angle. This is the angle between the direction of the sensor's laser beam, represented by $\overrightarrow{l}$, and the normal vector of the workpiece surface, $\overrightarrow{n}$. (d) Distance between two consecutive profiles $\Delta s$.}
    \label{fig:parametrosTrajectory}
    \end{figure}

One of these crucial parameters is the Optimal Working Distance ($W_d$), which denotes the ideal distance between the sensor and the object's surface. This distance ensures optimal precision of the captured data by positioning the laser source at the scanning reference plane, typically located at the midpoint of the depth of field.

The Depth of Field ($Z_r$) refers to the range of distances within which the sensor can accurately capture surface data during a single scan (Figure \ref{fig:parametrosTrajectory}(b)).  Assuming a point in the scanner's coordinate system is $(x_s, 0, z_s)$, the equation \ref{eq:zrange} must be satisfied. Operating within this range is critical as it minimizes noise levels associated with deviations from the optimal working distance. Studies \cite{Sadaoui2022} have demonstrated that maintaining proximity to the optimal working distance reduces noise, thereby enhancing measurement accuracy.
    
\begin{equation}
    W_d - \frac{Z_r}{2} \leq z_s \leq W_d + \frac{Z_r}{2}
    \label{eq:zrange}
\end{equation}

Another crucial parameter is the Direction Angle ($\alpha$), which signifies the angle between the sensor's orientation vector $\overrightarrow{l}$ and the normal vector $\overrightarrow{n}$ of the workpiece surface (Figure \ref{fig:parametrosTrajectory}(c). This angle is computed using Equation \ref{eq:angInc}.  As the direction angle increases, there is a higher likelihood of introducing noise into the capture. This phenomenon occurs because the scanner may capture unwanted reflections of the laser light and variations in surface reflectivity, negatively impacting the data quality. Previous studies \cite{Mahmud2011, Sadaoui2022} have empirically shown how noise levels correlate with the direction angle, highlighting its significance in achieving precise surface capture.

\begin{equation}
    \alpha = acos(-\overrightarrow{l} \cdot \overrightarrow{n})
    \label{eq:angInc}
\end{equation}

Additionally, the Distance Between Profiles ($\Delta s$) determines the density of points between consecutive scan profiles. Adequate point density ensures comprehensive coverage and accuracy of the inspected surface, particularly in areas with small features or irregular surfaces where a lower density might compromise inspection quality. See figure \ref{fig:parametrosTrajectory}(d).

In addition to considering these parameters, it is crucial to choose the appropriate type of trajectory to achieve a complete scan of the surface of the piece. In laser profilometer inspection applications, one of the most common strategies is to use Boustrophedon paths \cite{10.1007/978-1-4471-1273-0_32}, \cite{GALCERAN20131258}.

In a Boustrophedon scan, the sensor moves in a straight line along one axis until it reaches the edge of the surface to be inspected. Then, it shifts laterally a predetermined distance and changes direction to move in the opposite direction along the initial axis. This pattern of movements is repeated until the entire surface is covered. In the context of surface inspection, this method is highly efficient in ensuring that every area of the piece's surface is scanned without omissions, thereby maximizing coverage and inspection accuracy.

\begin{figure}
    \centering
    \includegraphics[width=0.5\columnwidth]{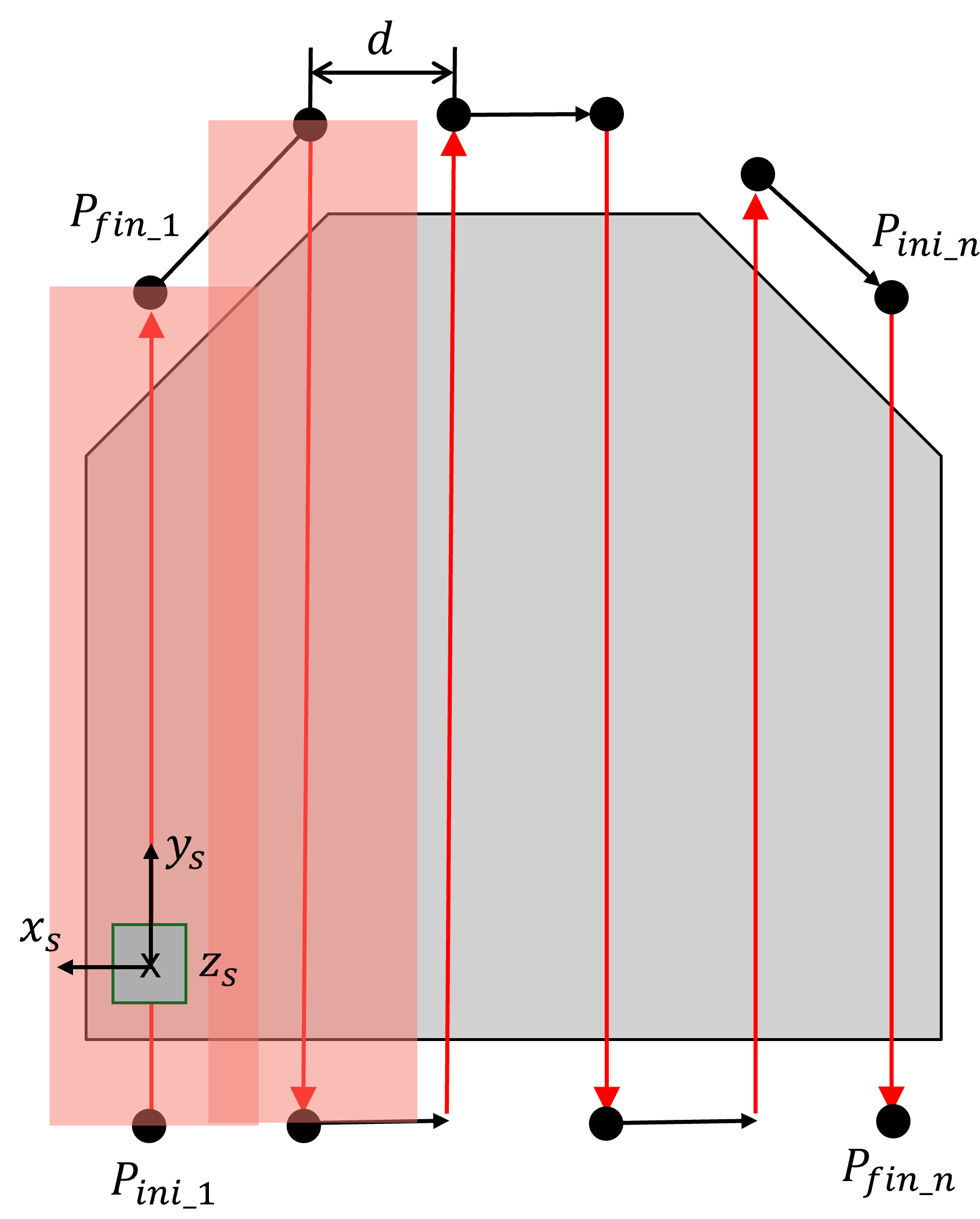}
    \caption{Top view of the scanning trajectory, where multiple parallel passes are made, each separated by a distance $d$. An overlapping area is defined between each pass. The gray square represents the sensor. The red lines show the trajectories where the sensor captures data. The black lines indicate the intermediate movements to position the sensor for the next pass.}
    \label{fig:Boustrophedon}   
\end{figure}

Considering these types of trajectories, the profilometric sensor collects data only during the parallel passes along the surface of the piece. In Figure \ref{fig:Boustrophedon}, these trajectories are shown in red, from the initial point of a pass ($P_{ini_i}$) to its end point ($P_{fin_i}$), where $i$ denotes the number of parallel passes. The movement of the robot between each pass is shown in black. The distance between passes, $d$, is carefully adjusted to ensure that the scans overlap adequately, thereby completely covering the piece.

\subsection{Simulated Environment}

To train the reinforcement learning algorithms, it is essential to have an environment that simulates the conditions of the real system. Conducting tests directly on the real system can be costly, dangerous, or simply impractical in many situations. Therefore, simulators are used to virtually recreate the environment of interest.

\begin{figure}
\centering
\includegraphics[width=1\columnwidth]{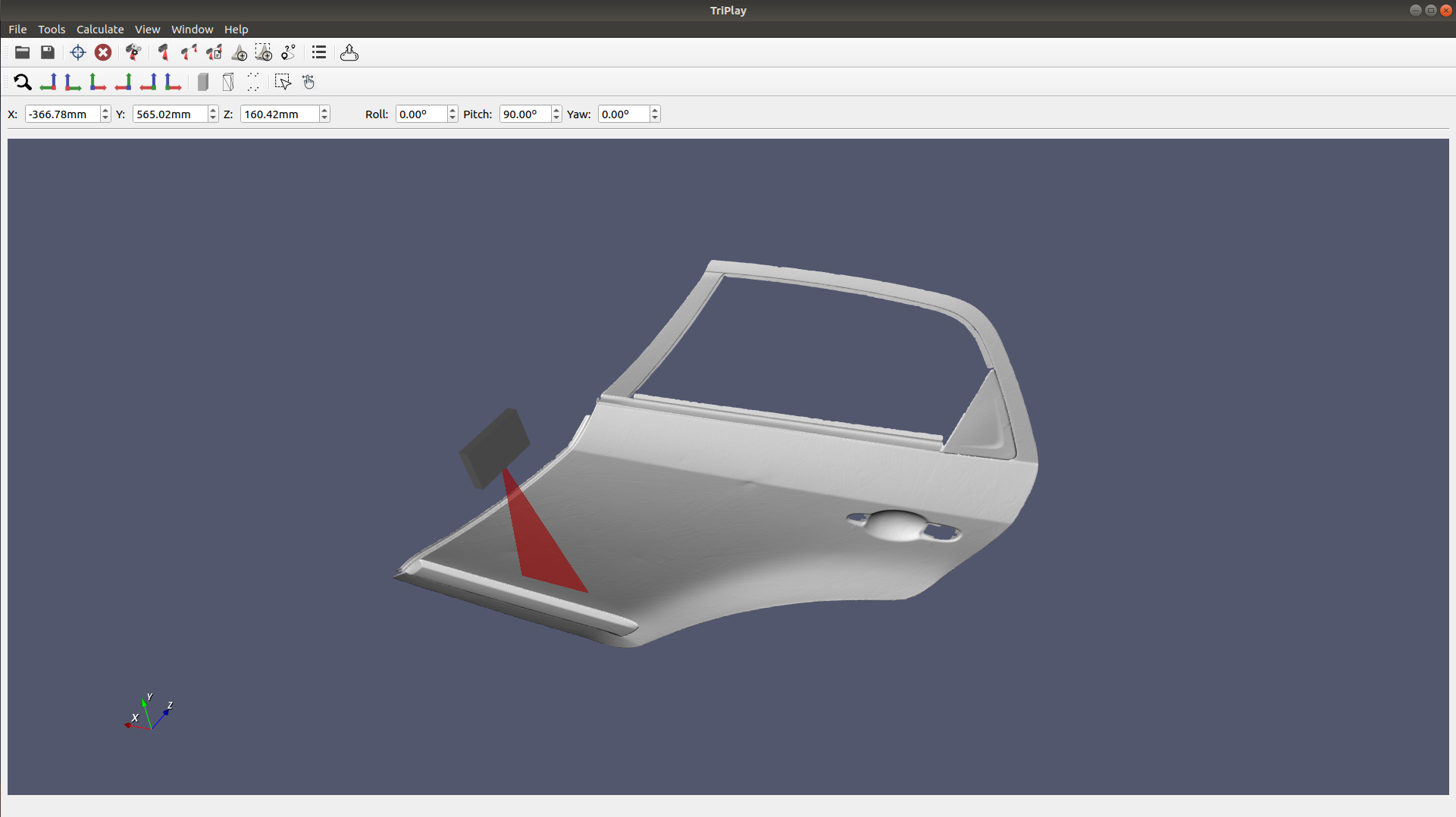}
\caption{View of the simulated environment with the profilometric sensor and a part to be inspected.}
\label{fig:simuCarDoor}
\end{figure}

We will use the simulator detailed in our previous work \cite{simu_roos}, designed to replicate the conditions of the real system in a virtual environment. This simulator recreates the measurements of a laser triangulation profilometric sensor, emulating the parameters of any commercial sensor according to its specification sheet. It allows for precise reproduction of measurements on a CAD model of the part to be inspected, including the introduction of inherent sensor noise and speckle noise generated by the object's surface. See Figure \ref{fig:simuCarDoor}.

\begin{figure}
\centering
\includegraphics[width=0.95\columnwidth]{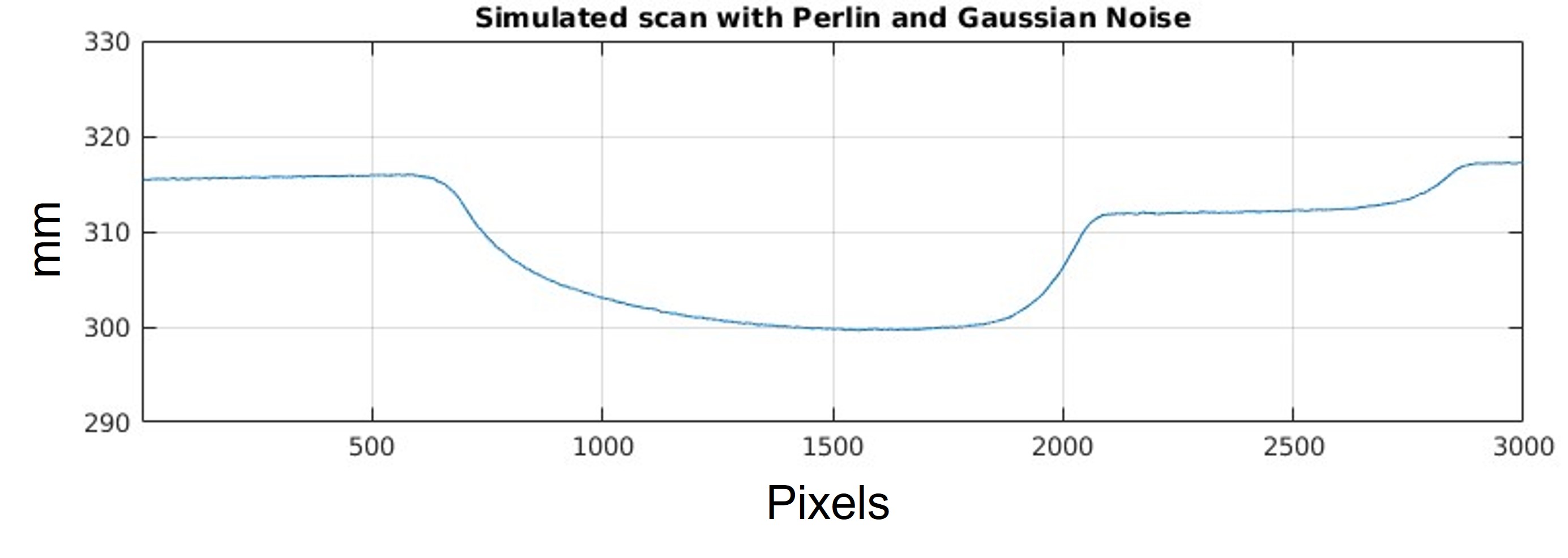}
\caption{Profile obtained during the simulation of a scan of the car door handle section of the CAD model shown in Figure \ref{fig:simuCarDoor}.}
\label{fig:Perfil}
\end{figure}

\begin{figure}
\centering
\includegraphics[width=0.8\columnwidth]{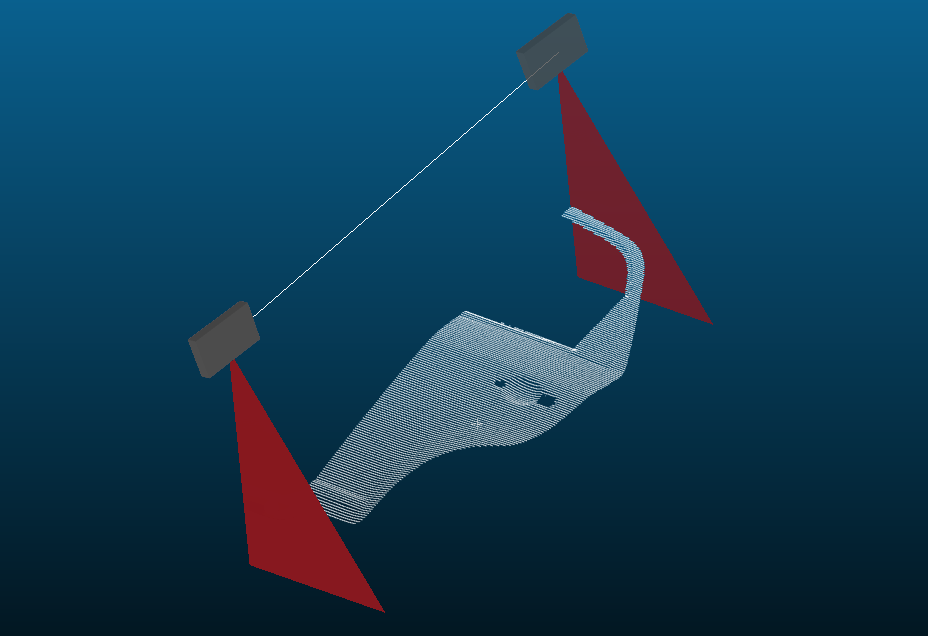}
\caption{Point cloud result obtained during the simulation of a scan of the CAD model section shown in Figure \ref{fig:simuCarDoor}. The start and end points of the trajectory can be seen.}
\label{fig:scan3D}
\end{figure}

In each iteration of the simulator, several critical parameters are obtained that will be used later by the RL algorithm. First, the distance profile is captured, a fundamental representation provided by any profilometric sensor, see Figure \ref{fig:Perfil}. Additionally, the 3D position of the scanned points of the CAD model is collected, providing detailed information about the surface geometry of the object, see Figure \ref{fig:scan3D}. Furthermore, the simulator also provides data on the normals at those points on the object's surface.

\subsection{State Space}

As previously mentioned the position and orientation of the end-effector are used instead of relying on the positions and velocities of the robot's joints. This choice simplifies the state space and facilitates the transfer of the method to different robotic configurations without the need for specific adjustments in the joints.

Mathematically, the state \( S \) is defined as a tuple as follows:

\begin{equation}
    S = \{P(x, y, z), \theta, D, \alpha, \Delta s\}
    \label{eq:stateSpace}
\end{equation}

Here, \(P(x,y,z)\) represents the position of the end-effector, while \(\theta\) denotes its tilt. The parameters \( D \), \( \alpha\), and \( \Delta s \) correspond to the mean profile distance obtained from the scan, the angle between the sensor and the surface, and the advance between consecutive scans in the 3D space, respectively.

\subsection{Action Space}

The action space is defined by the increments in the position and tilt angle of the inspection sensor. These increments are defined relative to the sensor's own coordinate system. Mathematically, the action space is represented by equation \ref{eq:actionSpace}.

\begin{equation}
A = \{\Delta y, \Delta z, \Delta \theta \}
\label{eq:actionSpace}
\end{equation}

\begin{figure}
    \centering
    \includegraphics[width=0.75\columnwidth]{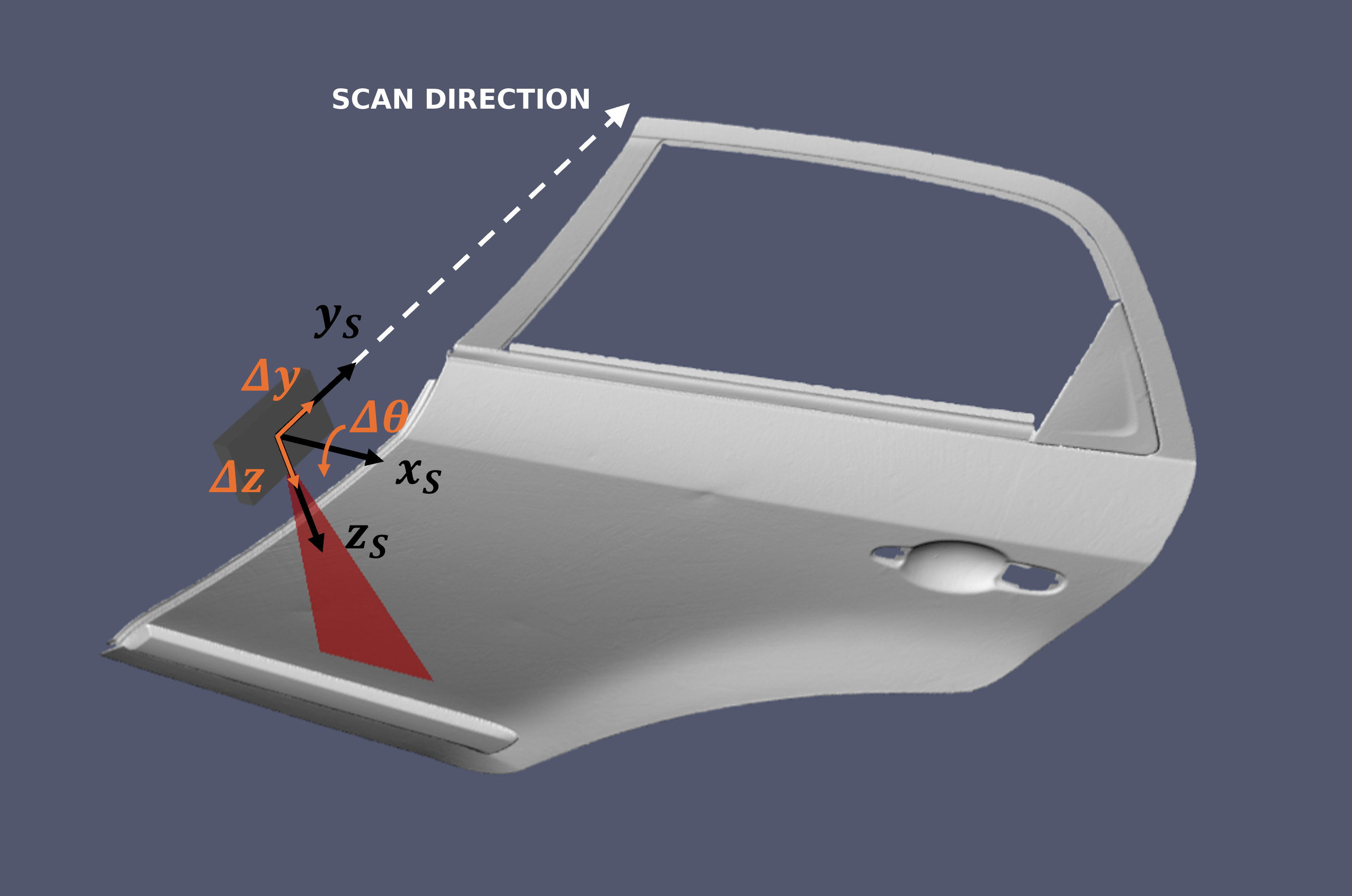}
    \caption{The figure shows the simulation environment and represents the action space as unit vectors (in orange). \(\Delta y\) refers to the increment in position in the scanning direction, \(\Delta z\) refers to the increment in the vertical direction (Z), and \(\Delta\theta\) denotes the change in the sensor's pitch orientation.}
    \label{fig:espacioAcciones}
\end{figure}

Where \(\Delta y\) refers to the increment in position in the sensor's forward direction (Y), which will be previously defined by a unit vector indicating the scanning direction. \(\Delta z\) refers to the increment in position in the sensor's vertical direction (Z), controlling the height of the end-effector relative to the part. \(\Delta\theta\) denotes the change in the sensor's pitch orientation, which is the rotation around the X-axis. This is represented in Figure \ref{fig:espacioAcciones}.

The action space is defined as continuous, meaning that actions span a continuous range of values rather than discrete ones. This approach ensures smooth and controlled sensor movements to avoid abrupt changes that could affect measurement accuracy or cause collisions with the workpiece. Equation \ref{eq:actionLimits} establishes the limits for each type of action in the continuous space. Here, $\Delta y$, $\Delta z$, and $\Delta \theta$ are constrained to values between $\pm \Delta y_{\max}$ millimeters, $\pm \Delta z_{\max}$ millimeters, and $\pm \Delta \theta_{\max}$ degrees, respectively.

\begin{equation}
 \begin{matrix}
   \Delta y \in [ - \Delta y_{max}, \Delta y_{max}] \\
   \Delta z \in [ - \Delta z_{max}, \Delta z_{max}] \\
   \Delta \theta \in [ - \Delta \theta_{max}, \Delta \theta_{max}]
 \end{matrix}
 \label{eq:actionLimits}
\end{equation}

\subsubsection{Dynamic Action Limitation}

To ensure smooth and safe movement of the inspection sensor, the selection of actions is dynamically adjusted based on the environment's observations. This action limitation accelerates the convergence of the reinforcement learning algorithm, enabling the system to learn more efficiently and effectively.

When the sensor is farther from the part surface than the optimal working distance \(W_d\), limits are applied to the sensor's displacement in the Z direction \(\Delta z\) to bring it closer to the surface in a controlled manner. Conversely, if the sensor is too close, displacements in the negative direction are limited, as per equation \ref{eq:clipActionD}.

\begin{equation}
\Delta z = \begin{cases}
    \text{clip}(\Delta z, 0, \Delta z_{max}) & \text{if } (D - W_d) \geq 0 \\
    \text{clip}(\Delta z, -\Delta z_{max}, 0) & \text{if } (D - W_d) < 0
\end{cases}
\label{eq:clipActionD}
\end{equation}

Here, \(clip(x,a,b)\) limits the value of \(x\) between \(a\) and \(b\), ensuring that the actions are within the permitted range, according to equation \ref{eq:clip}.

\begin{equation}
    \text{clip}(x,a,b) = \left\{\begin{matrix}
    a\ \text{if}\ x \leq a\\ 
    b\ \text{if}\ x \geq b\\ 
    x\ \text{else}
    \end{matrix}\right.
    \label{eq:clip}
\end{equation}

Similarly, if the sensor's direction angle (\(\alpha\)) with respect to the surface normal is positive, indicating excessive tilt, limits are applied to the angular displacement \(\Delta \theta\) to correct the sensor's orientation. Conversely, if the tilt angle is negative, limits are applied to the angular displacement in the opposite direction to keep the sensor properly aligned with the inspected surface. This is represented in equation \ref{eq:clipActionN}.

\begin{equation}
\Delta \theta = \begin{cases}
    \text{clip}(\Delta \theta, 0, \Delta \theta_{max}) & \text{if } \alpha \geq 0 \\
    \text{clip}(\Delta \theta, -\Delta \theta_{max}, 0) & \text{if } \alpha < 0
\end{cases}
\label{eq:clipActionN}
\end{equation}

\subsection{Reward Function}

In reinforcement learning, creating an effective reward model is crucial as it guides the agent toward desirable behaviors within the environment. This model assigns a value to each state-action pair, reflecting the immediate benefit or cost associated with the agent's decision. This section details the reward strategy designed in this research.

The proposed reward function \(R(s,a)\) consists of three distinct components, each capturing different aspects of the inspection process. Mathematically, this function is expressed as shown in equation \ref{eq:reward}.

\begin{equation}
    R(s,a)=w_d R_D+ w_{\alpha} R_\alpha+ w_{\Delta s} R_{\Delta s}
    \label{eq:reward}
\end{equation}

\(R_D\) represents the reward related to the distance between the sensor and the inspected surface, \(R_{\alpha}\) denotes the reward related to the alignment of the sensor's orientation with the normal of the inspected object's surface, and \(R_{\Delta s}\) captures the reward associated with the sensor's movement between consecutive scans in the 3D space corresponding to the point cloud of the inspected piece. \(w_d, w_{\alpha}, w_{\Delta s}\) represent the weights that each component contributes to the overall reward function.

The proposed rewards are in the range [0, -1], as the reward function aims to incentivize the agent to perform actions that improve the inspection process. The maximum value of 0 is assigned when the optimal goal is reached, while negative values indicate penalties for deviations from the desired behavior.

\subsubsection{Distance Reward \((R_D)\)}

To ensure that the sensor maintains an optimal distance from the inspected surface, a distance reward function \(R_D\) is defined as a continuous penalty function that decreases as the absolute difference between the observed distance and the optimal working distance \(W_d\) increases. The reward function is formulated as follows:

\begin{equation}
R_D = - \frac{(W_d - D)^2}{(\frac{Z_{r}}{2})^2}
\label{eq:RewD}
\end{equation}

\begin{figure}
    \centering
    \includegraphics[width=0.7\columnwidth]{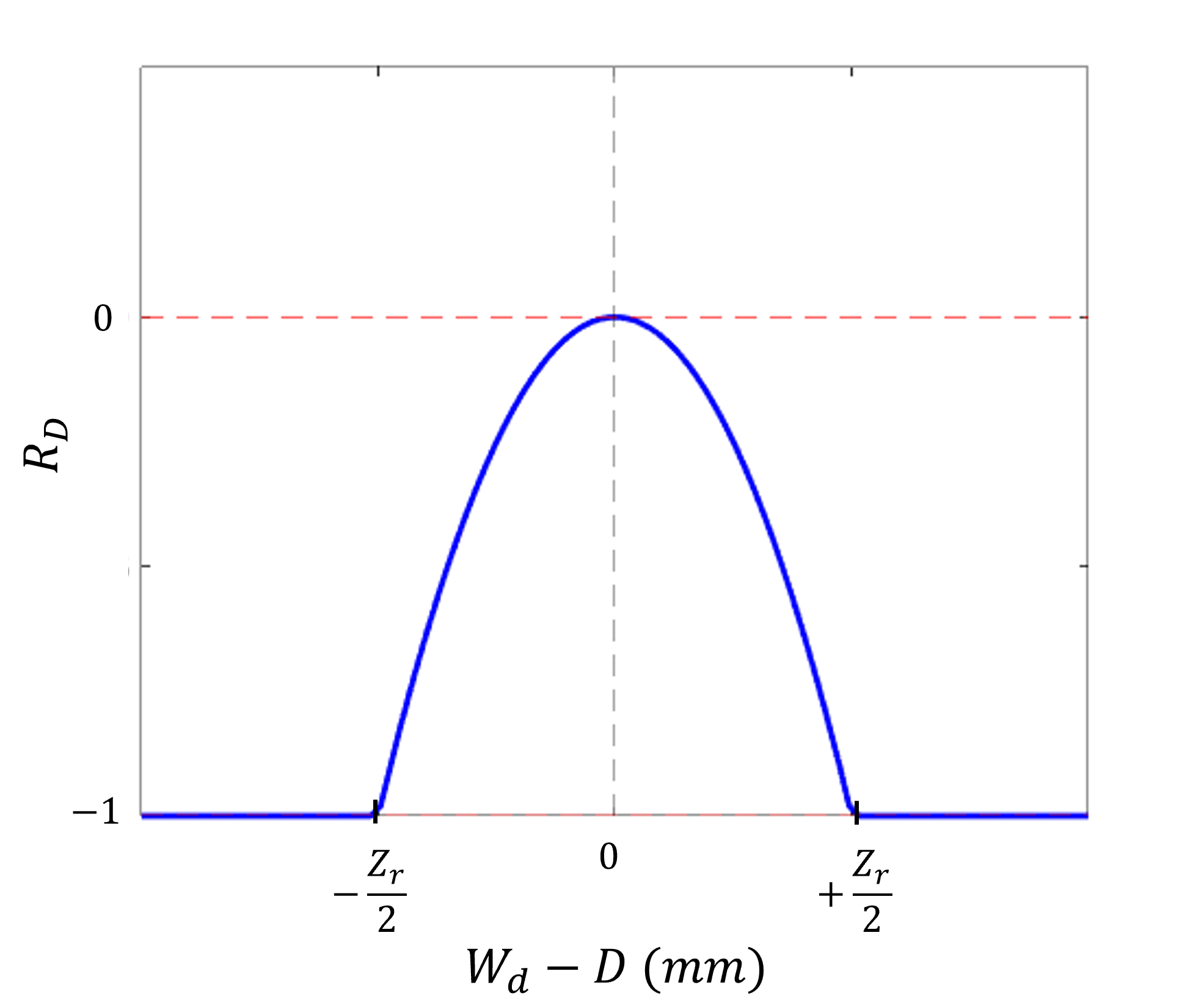}
    \caption{Graph of the Distance Reward Function \(R_D\). The graph shows the relationship between the observed distance \(D\) and the reward \(R_D\) based on the difference from the optimal working distance \(W_d\).}
    \label{fig:RewardDGraphic}
\end{figure}

Where \(W_d\) represents the optimal working distance, \(D\) the observed distance during scanning, and \(Z_{r}\) the specified working range of the sensor. This results in a parabolic function with values between [-1,0], corresponding to 0 when operating at the optimal working distance and -1 at the sensor's range limits, as shown in Figure \ref{fig:RewardDGraphic}. If the distance is outside this range, the penalty is maximum (-1).

\subsubsection{Orientation Reward \((R_\alpha)\)}

\begin{figure}
    \centering
    \includegraphics[width=0.7\columnwidth]{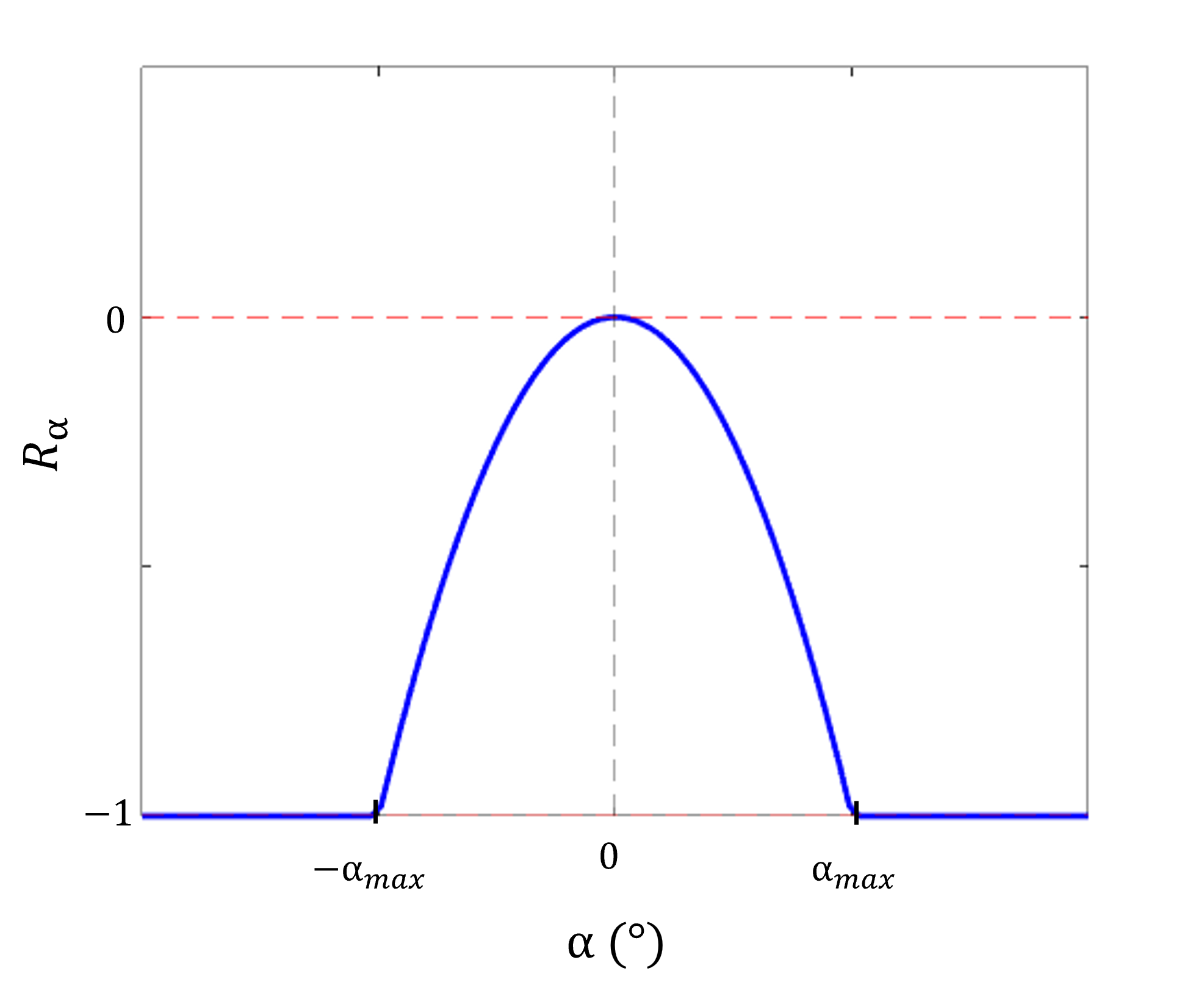}
    \caption{Graph of the Orientation Reward Function \(R_\alpha\). The graph shows the relationship between the direction angle\(\alpha\) and the reward \(R_\alpha\) according to a maximum direction angle \(\alpha_{max}\).}
    \label{fig:RewardAlphaGraphic}
\end{figure}

To induce the agent to align its orientation with the surface normal, we introduce an orientation reward model ($R_{alpha}$). This model is designed to minimize the angular disparity between the sensor direction and the surface normal vector. The function is defined as a continuous penalty function that approaches 0 as the absolute orientation difference decreases, see Figure \ref{fig:RewardAlphaGraphic}:

\begin{equation}
    R_\alpha = \max(-1, - \frac{\alpha^2}{\alpha_{max}^2})
    \label{eq:RewN}
\end{equation}

Where \(\alpha\) is the angular difference between the sensor's orientation and the surface normal, and \(\alpha_{max}\) is the maximum allowed angular disparity threshold. This model encourages the agent to maintain close alignment with the surface normal, optimizing the quality of the inspection.

\subsubsection{Movement Reward \((R_{\Delta s})\)}

In addition to optimizing the distance and orientation of the sensor, ensuring smooth forward movement is crucial for comprehensive surface coverage. Forward scanning movement ensures that each scanned 3D profile extends beyond the previous one, facilitating thorough inspection. The reward function \(R_{\Delta s}\) is expressed as:

\begin{equation}
    R_{\Delta s} = \max(-1, - \frac{(\Delta s - \Delta s_{opt})^2}{\Delta s_{opt}^2})
        \label{eq:ReA}
\end{equation}

\begin{figure}
    \centering
    \includegraphics[width=0.7\columnwidth]{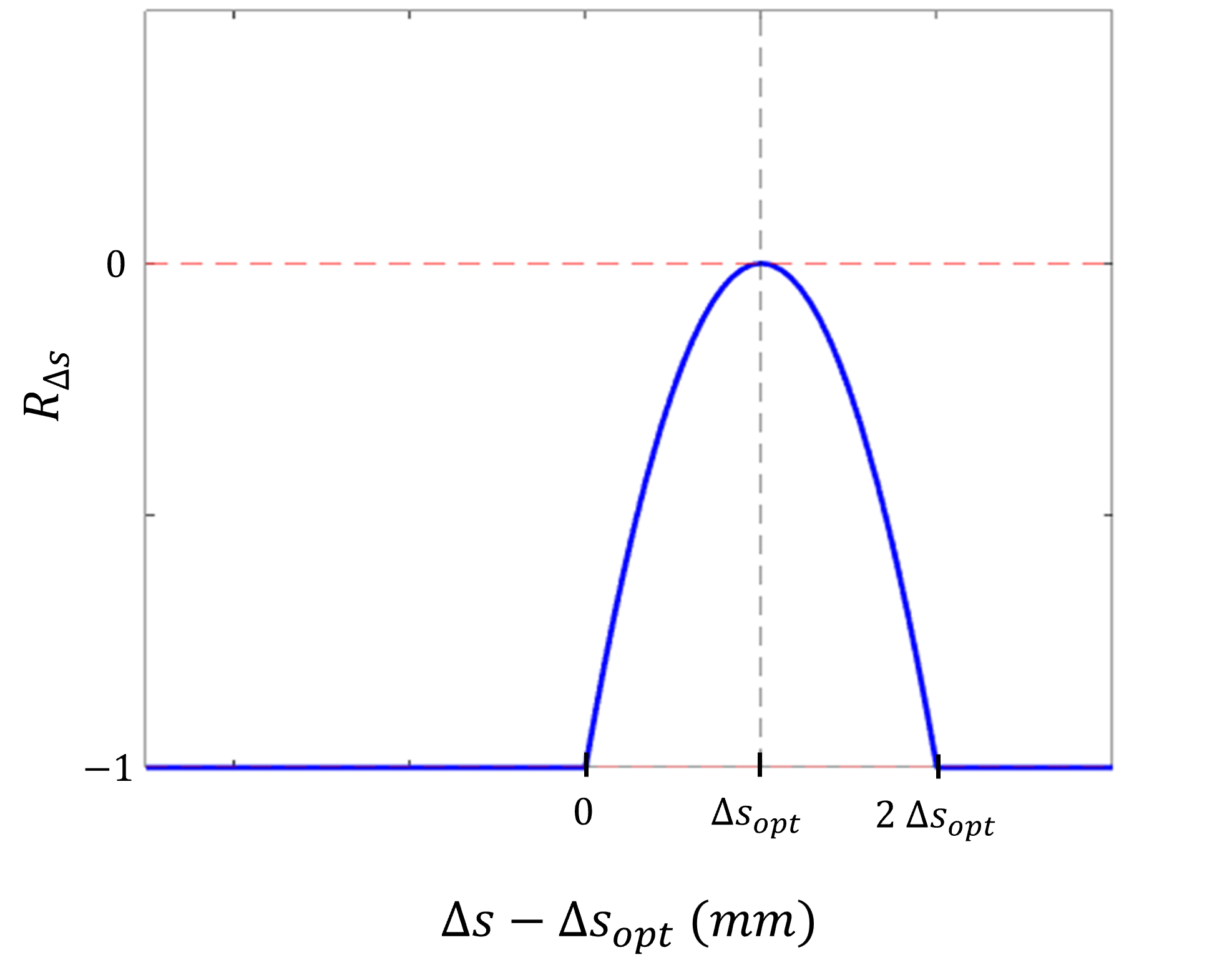}
    \caption{Graph of the Movement Reward Function \(R_{\Delta s}\). The graph shows the relationship between the scanning spacing between the current and previous profile \(\Delta s\) and the reward \(R_{\Delta s}\) according to an optimal spacing \(\Delta s_{opt}\).}
    \label{fig:RewardDeltaSGraphic}
\end{figure}

This function penalizes the agent when the scanning spacing \(\Delta s\) is negative, indicating backward movement within the inspection area. Likewise, it behaves parabolically with respect to the scanning spacing \(\Delta s\). When the spacing is equal to twice the optimal value \(\Delta s_{opt}\), the reward reaches its minimum value of -1. This indicates a strong penalty for excessively large spacings. As the spacing decreases from this point, the reward gradually increases, reaching a maximum value of 0 when the spacing is exactly equal to the optimal value. Therefore, the reward function motivates the agent to maintain spacing close to the optimal, as both above and below-optimal values result in a decrease in reward, see Figure \ref{fig:RewardDeltaSGraphic}.

\subsection{RL Algorithm: PPO}

The Proximal Policy Optimization (PPO) algorithm \cite{SchulmanWDRK17} is a policy gradient technique designed to provide faster and more efficient updates than previously developed reinforcement learning algorithms, such as Advantage Actor-Critic (A2C) or Deterministic Policy Gradient (DPG).

PPO was designed as an improvement over Trust Region Policy Optimization (TRPO). It simplifies and accelerates the training process by using first-order gradients and a clipped objective function that stabilizes policy updates.

PPO employs a clipped surrogate loss function, penalizing excessive changes in the policy, stabilizing training, and preventing significant divergence between new and old policies. The clipped objective function of PPO is defined by equation \ref{eq:PPO}.

\begin{equation}
    J_{\text{clip}}(\pi) = \mathbb{E} \left[ \min \left( r(\pi) \hat{A}, \text{clip}\left( r(\pi), 1 - \epsilon, 1 + \epsilon \right) \hat{A} \right) \right]
    \label{eq:PPO}
\end{equation}

Here, $\pi$ represents the policy, $\mathbb{E}$ denotes the expectation over time, $r$ is the ratio of probability under the new and old policies, respectively, $\hat{A}$ is the estimated advantage, and $\epsilon$ is a hyperparameter controlling how much the new policies are allowed to differ from the old policies during the optimization process. It is used to compute a penalty function that limits the policy change at each optimization iteration. The probability ratio $r(\pi)$ is calculated as the probability of selecting an action under the new policy divided by the probability of selecting the same action under the old policy, i.e.:

\begin{equation}
    r(\pi) = \frac{\pi_{\text{new}}(a_t|s_t)}{\pi_{\text{old}}(a_t|s_t)}
\end{equation}

For the PPO algorithm, the hyperparameter configuration involves several key aspects. The neural network architecture defines the structure of the neural network, including the number of hidden layers and units per layer. An activation function, typically ReLU, is applied to the outputs of each hidden layer to introduce non-linearities into the model. The learning rate determines the size of the step taken during weight updates, influencing the speed and stability of learning. Additionally, the clip ratio limits policy changes between updates to ensure training stability. The epoch parameter denotes the number of complete passes through the training dataset during training. 

\section{Results}

In this section, we present the experiments conducted to validate and evaluate the proposed methods in the context of generating inspection trajectories using reinforcement learning (RL) algorithms. These algorithms were implemented using the open-source library stable-baselines3 \cite{stable-baselines3}, which provides enhanced implementations of reinforcement learning algorithms based on OpenAI. To analyze and process the obtained results, we used MATLAB 2023b.

First, we detail the training process of the RL model, including the architecture of the neural network used, the configured hyperparameters, and the training methodology employed. 

\begin{figure}
    \centering
    \subfloat[]{
    \includegraphics[width=0.33\columnwidth]{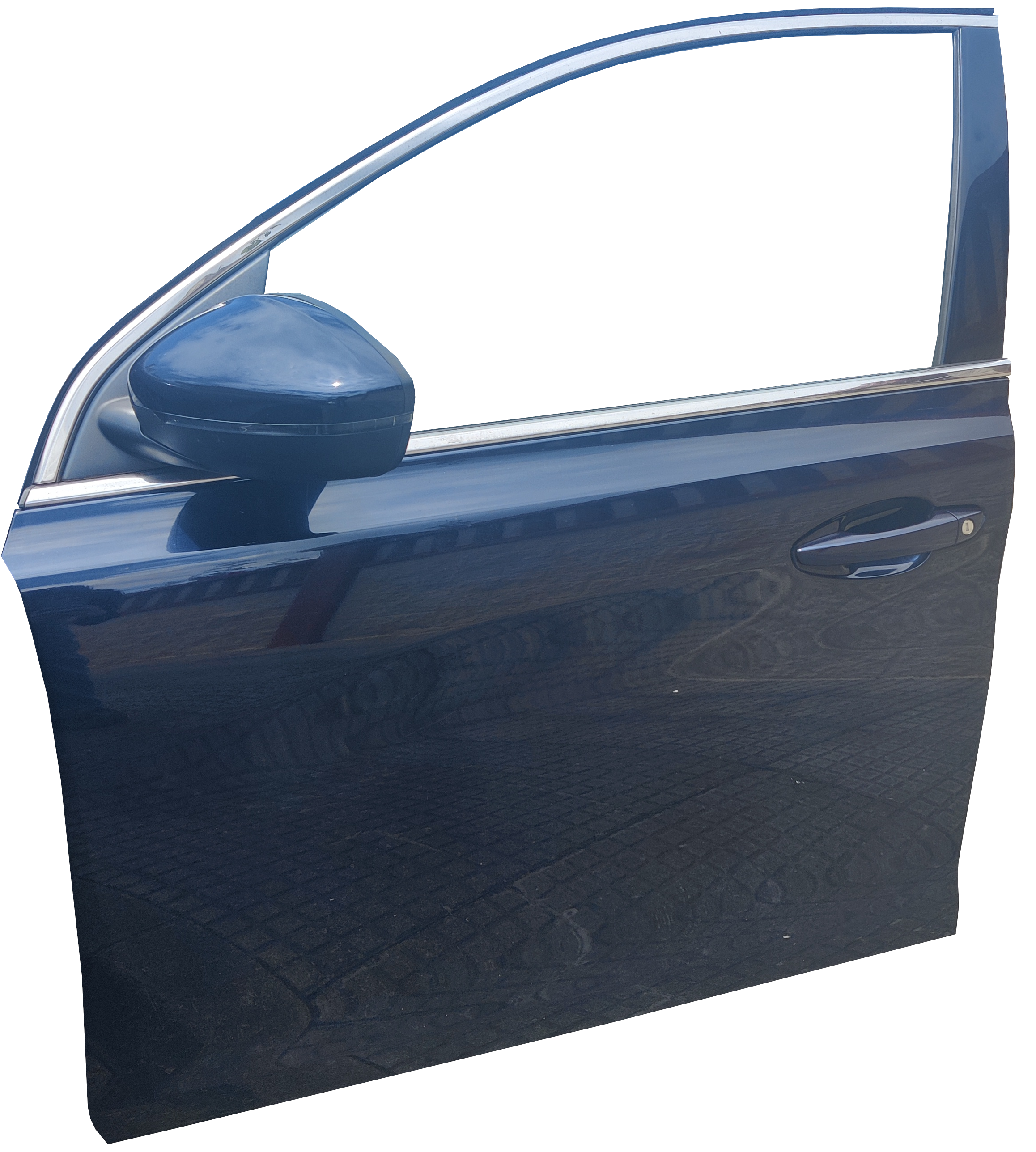}}
         \subfloat[]{
    \includegraphics[width=0.55\columnwidth]{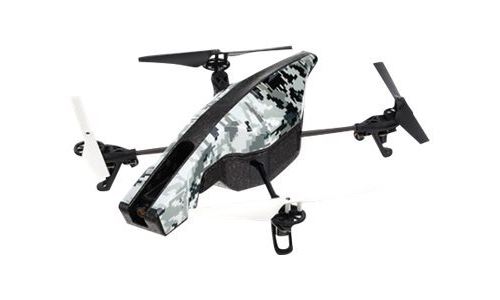}}
    \caption{Parts used for the experiments. (a) Car door. (b) Parrot Drone.}
    \label{fig:piezasExperimentos}
\end{figure}

Subsequently, the trained RL model is employed to generate inspection trajectories for two different parts using their CAD models: a car door and the body of a Parrot drone, see figure \ref{fig:piezasExperimentos}. Due to its dimensions, the car door will be scanned using a Boustrophedon trajectory base, whereas the drone will be scanned with a single straight-line scan. 

The trained RL model takes the CAD model of the part as input and produces a sequence of movements that the sensor must follow to efficiently scan its surface. This trajectory is designed to minimize the error between the actual distance of the sensor to the part and the sensor's optimal working distance, as well as the direction angle. Additionally, it ensures a constant separation between scan profiles, guaranteeing uniform and precise coverage of the entire surface.

We perform different experiments, both in simulation and in a real environment. Simulation results obtained during the execution of the trajectories in our inspection simulator are presented and analyze. The results of the scans generated by RL optimized trajectories are compared with conventional methods such as straight line trajectories or Boustrophedon type trajectories. 

Finally, an experiment is conducted using a UR3e robot in a real-world environment to execute the inspection trajectory generated offline by the trained RL model for the Parrot drone. We analyze the results obtained to validate the transferability of the solution from the simulated environment to practical real-world applications.

\subsection{Training Process}

The training process of the RL model for trajectory optimization in robotic inspection was developed using a detailed simulation environment, the characteristics of which are explained in \cite{simu_roos}. In this context, a profilometric sensor was simulated with the same specifications as the Automation Technology model AT-C5-4090CS30-495, whose main parameters are detailed in Table \ref{tab:sensorParams}. It is important to note that this setup is designed to generalize based on input parameters, allowing for adjustments to different working distance, for example.

\begin{table}
    \centering
    \caption{Parameters of the profilometric sensor extracted from its datasheet and the velocity used in these experiments.}
    \begin{tabular}{ | c | c | }
    \hline
    \textbf{Parameters} & \textbf{Value} \\
    \hline
    Working Distance & 400 mm \\ 
    Z Range & 250 mm \\ 
    Field of View (FOV) & 63.5$^{\circ}$ \\ 
    Points per Profile & 4096 pixels \\ 
    Z Resolution & 3.8 $\mu$m \\
    \hline
    \end{tabular}
    \label{tab:sensorParams}
\end{table}

\begin{figure}
\centering
\includegraphics[width=0.9\columnwidth]{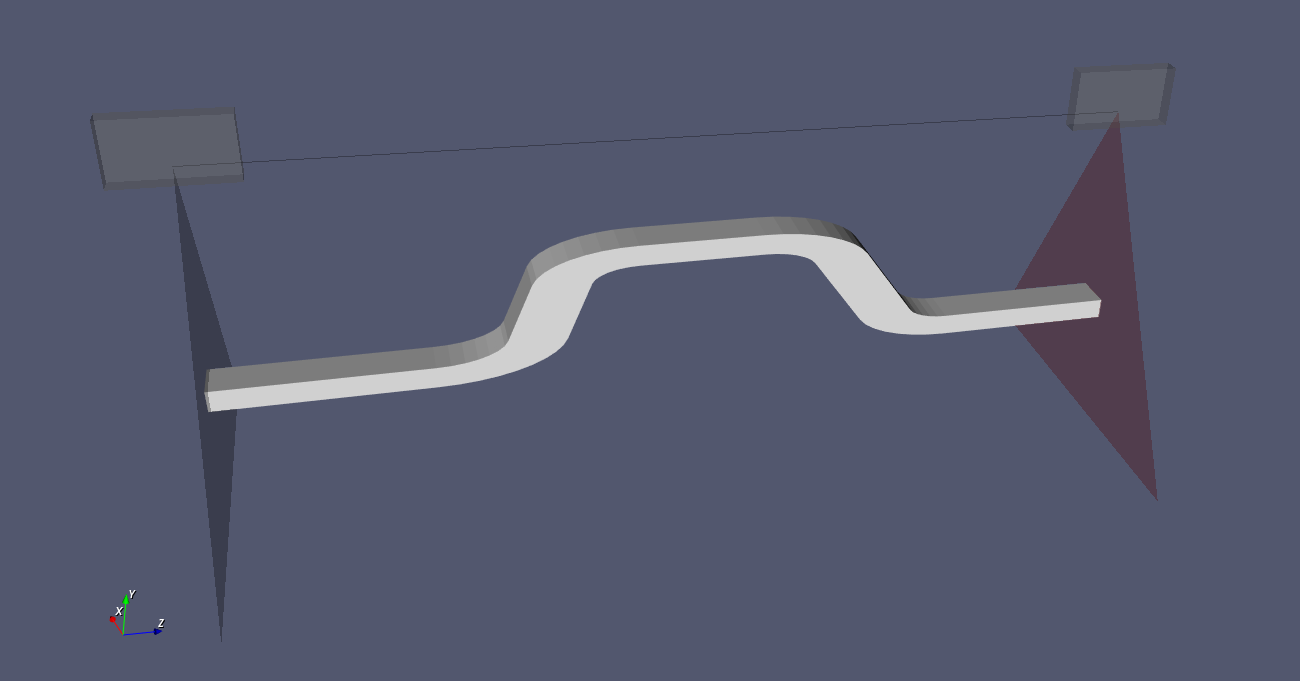}
\caption{Environment used for the reinforcement learning model training. The CAD model of the piece used and the start and end poses of the trajectory to be optimized are shown.}
\label{fig:trainingPiece}
\end{figure}

The design of the training piece was aimed at representing a wide variety of conditions that could be encountered in real inspection applications. This piece, created in 3D modeling software, features changes in orientation, height variations, and flat surfaces. Its dimensions are 1050x150x50mm, as shown in Figure \ref{fig:trainingPiece}.

During training, each episode is defined so that it corresponds to a starting point and an ending point, determined by the scanning direction and the piece's dimensions, visually represented in the same figure showing the CAD model of the piece used for training, as illustrated in Figure \ref{fig:trainingPiece}.

In the experiments, the action space is continuous, meaning actions are expressed as values within a continuous range rather than discrete values. Specifically, these actions are limited within the interval of [-1, 1], where position increments are measured in millimeters and pitch angles in degrees.

Table \ref{tab:paramsPPO} summarizes the hyperparameters used for the PPO algorithm. These parameters were selected based on default values recommended by the authors of the stable-baselines3 library. The neural network architecture consists of two hidden layers with 64 units each and ReLU activation function. A learning rate of 0.0003 was employed, with updates performed every 2048 steps and a batch size of 64. The discount factor ($\gamma$) was set to 0.99, and a clip ratio of 0.2 was used to stabilize training by limiting policy updates. Training proceeded over 10 epochs to refine the policy effectively. These hyperparameters were chosen to balance exploration and exploitation, ensuring robust and efficient learning within the RL framework.

\begin{table}
    \centering
    \caption{Hyperparameters of the PPO algorithm.}
    \begin{tabular}{|c|c|}
    \hline
    \textbf{Hyperparameters} & \textbf{Value} \\
    \hline
    Neural Network Architecture & [64,64] \\
    Activation Function & ReLU \\
    Learning Rate & 0.0003 \\
    Update Rate & 2048 steps \\
    Batch Size & 64 \\
    Discount Factor ($\gamma$) & 0.99 \\
    Clip Ratio & 0.2 \\
    Epoch & 10 \\
    \hline
    \end{tabular}
    \label{tab:paramsPPO}
\end{table}

During the training process of the algorithm, various metrics are employed to assess their performance and convergence capability. These metrics include the reward per episode, the length of episodes, and the number of episodes required to reach a certain performance level. The reward per episode is a crucial metric indicating the total amount of reward accumulated by the agent in each training episode. Generally, a higher reward reflects better performance of the agent in the task.

However, in this specific training context, evaluating solely the accumulated reward per episode might not be the most appropriate approach. This is because the length of episodes can vary significantly depending on the step size, defined as the distance between profiles. Therefore, instead of focusing solely on the accumulated reward, it is preferred to evaluate the globally normalized reward by the length of the episode. This metric provides a more comprehensive assessment of the agent's performance, as it considers both the accuracy of measurements and the efficiency of the trajectory. By doing so, a more precise insight into the overall effectiveness of the model in trajectory optimization and inspection quality is obtained, regardless of the specific length of episodes.

Additionally, the total number of episodes required for the agent to reach a certain performance level or convergence is analyzed. This number gives us an idea of how quickly the RL algorithm can learn and improve its performance in the given task.

 \begin{figure}
    \centering
    \subfloat[]{
        \includegraphics[width=0.5\columnwidth]{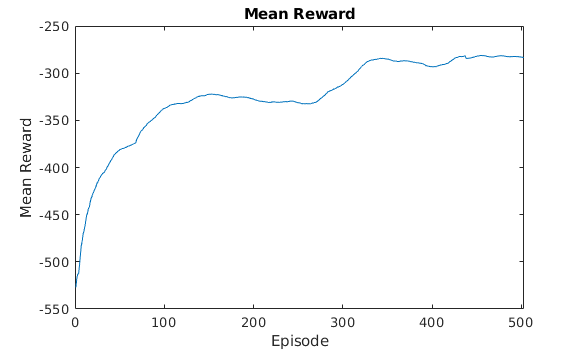}} 
    \subfloat[]{
             \includegraphics[width=0.5\columnwidth]{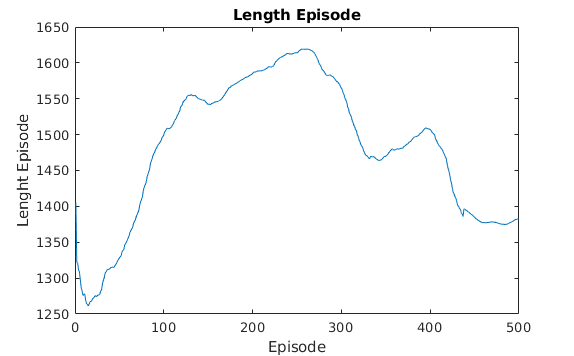}} \\
    \subfloat[]{
        \includegraphics[width=0.5\columnwidth]{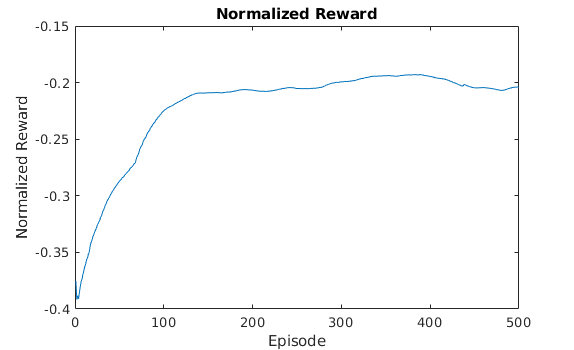}}
    \caption{Training metrics: (a) mean episodic reward, (b) mean episode length, and (c) normalized reward over the course of training.}
    \label{fig:trainingMetrics}
\end{figure}

In this experiment, the weight of each partial reward has been set so that they have the same influence on the overall reward, that is, $w_D = w_\alpha = w_{\Delta s} = 1/3$.


 \begin{figure}
    \centering
        \includegraphics[width=0.9\columnwidth]{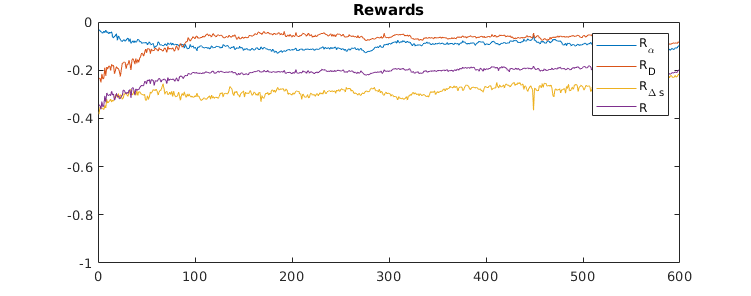}
    \caption{Comparison of overall reward ($R$) and partial rewards ($R_D, R_\alpha, R_{\Delta s}$) during training}
    \label{fig:trainingRewards}
\end{figure}


\subsection{Car door}

\begin{figure}
    \centering
    \includegraphics[width=0.5\columnwidth]{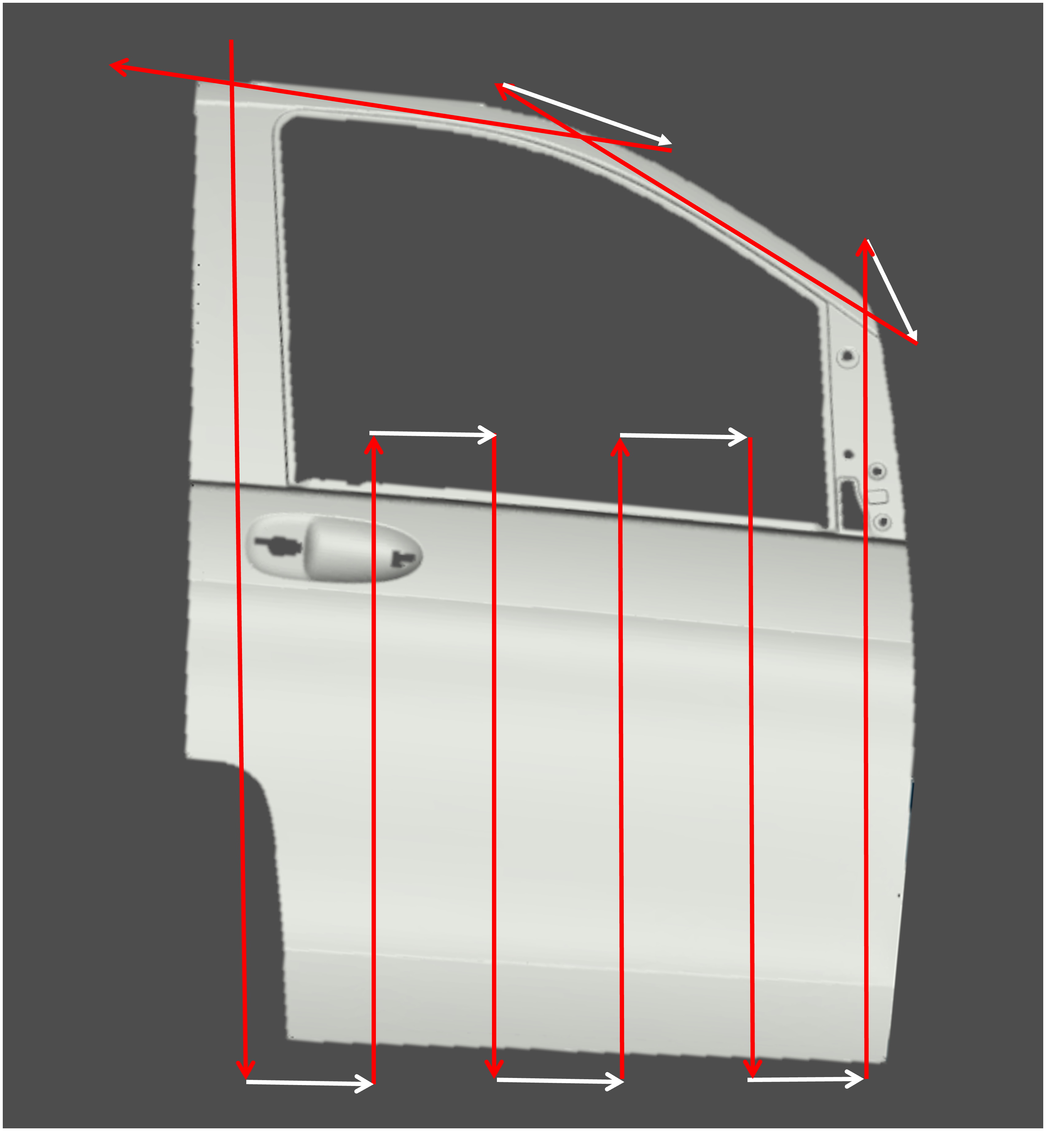}
    \caption{Initial Boustrophedon-type trajectory for door scanning. Inspection passes to be optimized are represented in red. White indicates movements between passes.}
    \label{fig:BoustPuerta}
\end{figure}

The first piece used to evaluate the model was a car door with dimensions 1440x1060x190mm. The initial trajectory, serving as a base for the optimization, is depicted in Figure \ref{fig:BoustPuerta}. In this figure, the initial Boustrophedon-type trajectory is observed, with inspection passes marked in red and movements between passes in white.

The reinforcement learning model is applied throughout the different passes, dynamically adjusting the orientation and distance of the profilometric sensor. Each pass of the Boustrophedon scan is optimized to maintain an appropriate pose between the sensor and the door surface, ensuring precise and uniform data capture. The model is exclusively applied to the red passes, where inspection is performed, optimizing the sensor's orientation and distance in each of these passes. The white movements represent transitions between passes where the sensor is not scanning and, therefore, not optimized.

Here are the results obtained when applying the trained model to the trajectory defined in Figure \ref{fig:BoustPuerta}. The definition of the start and end points of each pass is manually performed using the simulator interface.

\begin{figure}
    \centering
    \subfloat[]{
    \includegraphics[width=0.5\columnwidth]{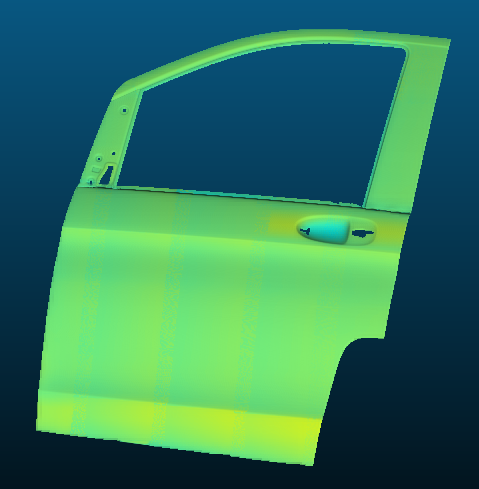}}
         \subfloat[]{
             \includegraphics[width=0.492\columnwidth]{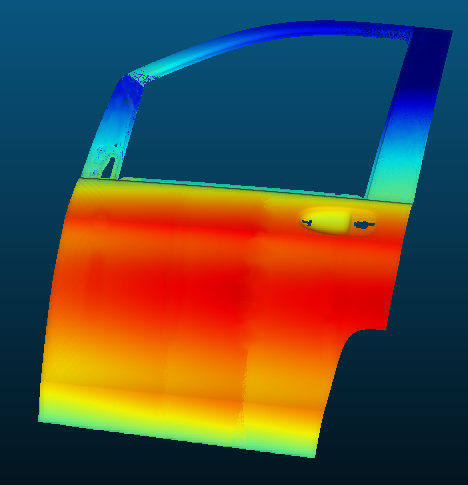}}
     \subfloat[]{
    \includegraphics[width=0.067\columnwidth]{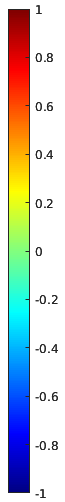}}
    \caption{Distance error map obtained during the scanning with profilometric sensor. The colors indicate the difference between the measured distance and the optimal sensor distance, normalized based on defined distance values. (a) Using trajectories that adapt to the surface of the piece, calculated by the RL algorithm. (b) Using straight trajectories defined by a start point and an end point.}
    \label{fig:mapDistancePuerta}
\end{figure}

Figure \ref{fig:mapDistancePuerta} (a) displays the point cloud obtained during the scanning using the profilometric sensor. The point cloud is depicted with a color map indicating the error in the measured distance for each point. This error is defined as the difference between the optimal working distance of the sensor and the actual distance obtained during the measurement at each point.

The aim of the analysis is to evaluate the accuracy and consistency of the profilometric sensor under real scanning conditions. The use of the color map facilitates the visualization of areas with higher and lower errors, providing a clear insight into the spatial distribution of the error across the scanned surface.

To provide a meaningful comparison, a scan was also carried out using a more traditional method. In this approach, the scanning was performed following the boustrophedon trajectory with a static configuration of height and orientation, without dynamic adjustments. This method is commonly employed in industrial applications due to its simplicity and ease of implementation. Figure \ref{fig:mapDistancePuerta} (b) shows the representation of the distance error map in a traditional scan.

\begin{figure}
    \centering
    \subfloat[]{
    \includegraphics[width=0.5\columnwidth]{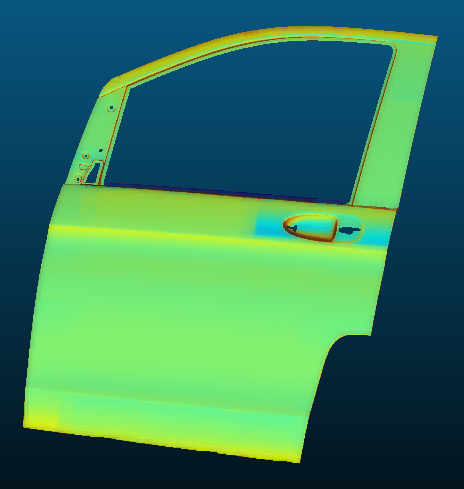}}
         \subfloat[]{
             \includegraphics[width=0.485\columnwidth]{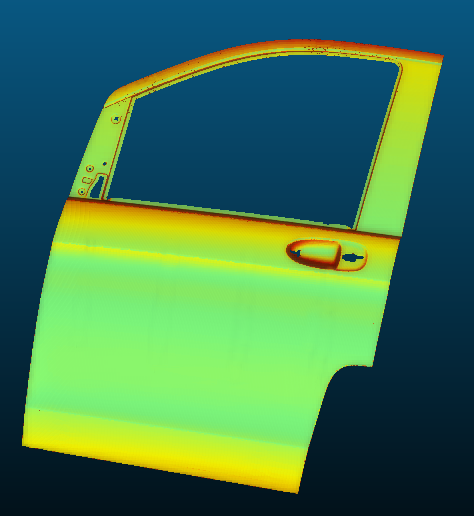}}
              \subfloat[]{
    \includegraphics[width=0.0687\columnwidth]{figuras/colorbar.png}}
    \caption{Orientation error map showing angular deviations from the optimal sensor orientation at each point, normalized based on defined orientation values. (a) Using trajectories that adapt to the surface of the piece, calculated by the RL algorithm. (b) Using straight trajectories defined by a start point and an end point.}
    \label{fig:mapOrientacionPuerta}
\end{figure}

In addition to the distance error map, an orientation error map was generated, which displays angular deviations from the optimal sensor orientation at each point. This deviation refers to the direction angle of the sensor on the surface. See Figure \ref{fig:mapOrientacionPuerta}.


\subsection{Parrot drone}


The results obtained from the optimization of the inspection trajectory of the Parrot drone are presented below. Figure \ref{fig:droneParrot} shows an image of the drone's body and its CAD model. The dimensions of the part are 350x95x65mm.

\begin{figure}
\centering
\includegraphics[width=0.9\columnwidth]{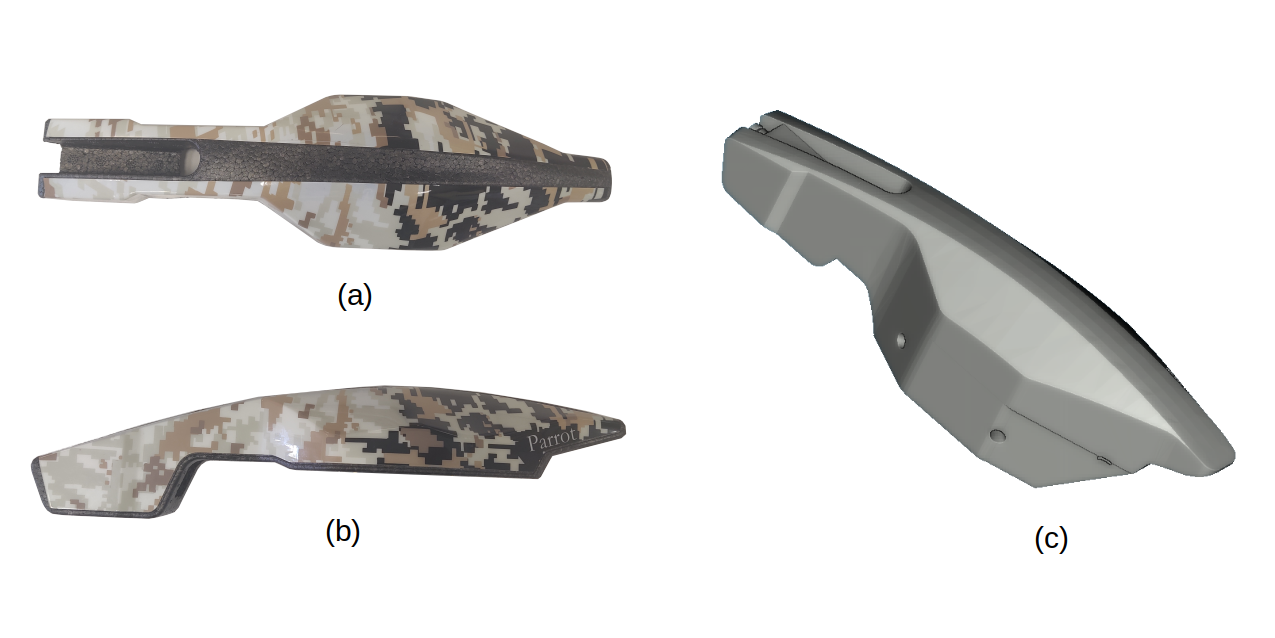}
\caption{Image of the Parrot AR. 2.0 drone body and its CAD model.}
\label{fig:droneParrot}
\end{figure}

\begin{figure}
\centering
\includegraphics[width=0.9\columnwidth]{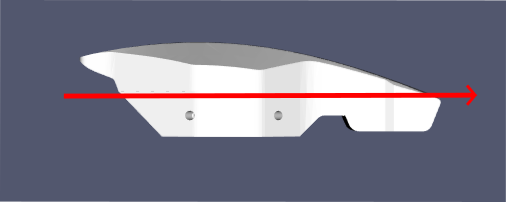}
\caption{Straight path followed during scanning.}
\label{fig:trajDrone}
\end{figure}

A scan of one side of the drone body will be performed. Due to its dimensions, the base trajectory used will be a simple straight line from the beginning of the part to the end, see figure \ref{fig:trajDrone}.

Using the trained reinforcement learning (RL) model, the optimized scanning trajectory for the drone is generated.


\begin{figure}
\centering
\subfloat[]{
\includegraphics[width=0.8\columnwidth]{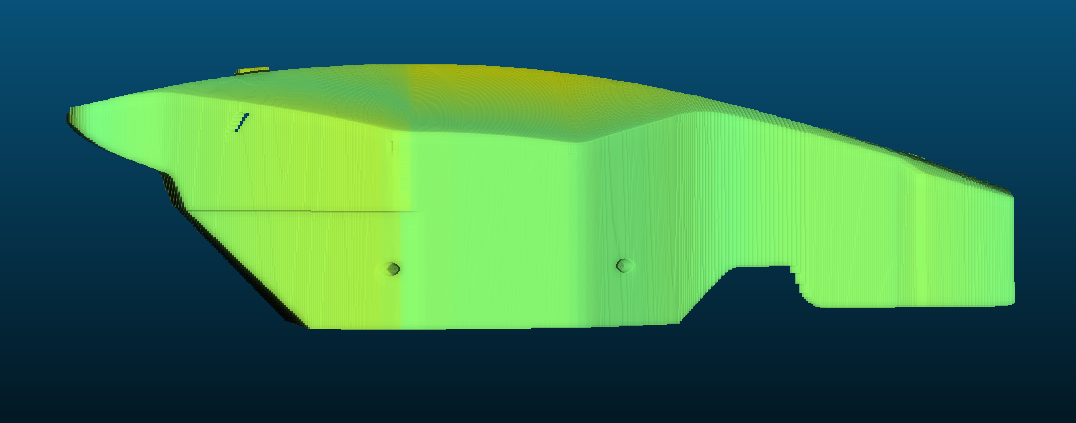}
\includegraphics[width=0.0415\columnwidth]{figuras/colorbar.png}}
\
\subfloat[]{
\includegraphics[width=0.8\columnwidth]{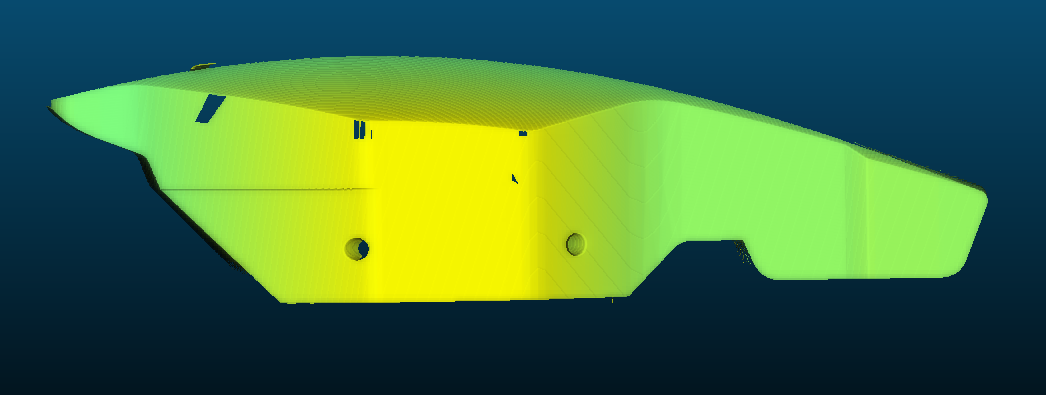}
\includegraphics[width=0.042\columnwidth]{figuras/colorbar.png}}
\caption{Distance error map obtained during the scanning with profilometric sensor. The colors indicate the difference between the measured distance and the optimal sensor distance, normalized based on defined distance values. (a) Using trajectory that adapt to the surface of the piece, calculated by the RL algorithm. (b) Using straight trajectory defined by a start point and an end point.}
\label{fig:expDroneSimu}
\end{figure}

\begin{figure}
\centering
\subfloat[]{
\includegraphics[width=0.8\columnwidth]{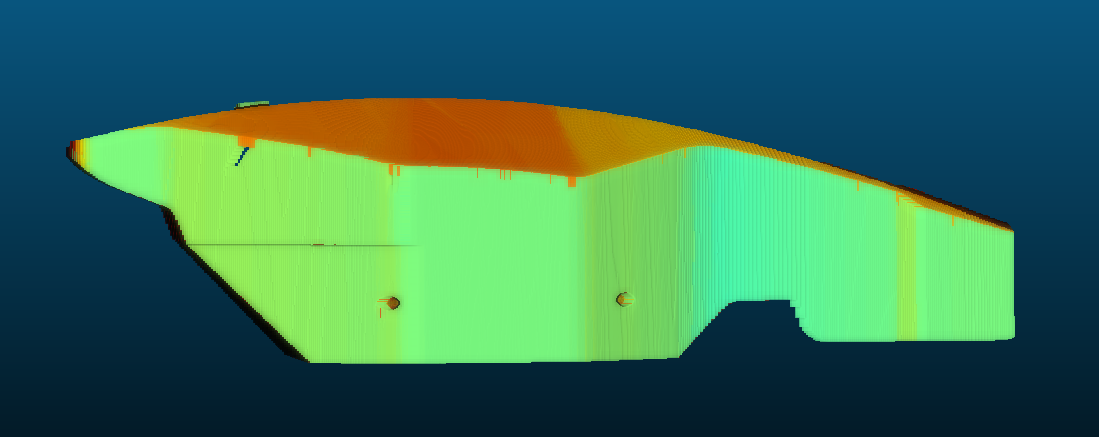}
\includegraphics[width=0.0415\columnwidth]{figuras/colorbar.png}}
\
\subfloat[]{
\includegraphics[width=0.8\columnwidth]{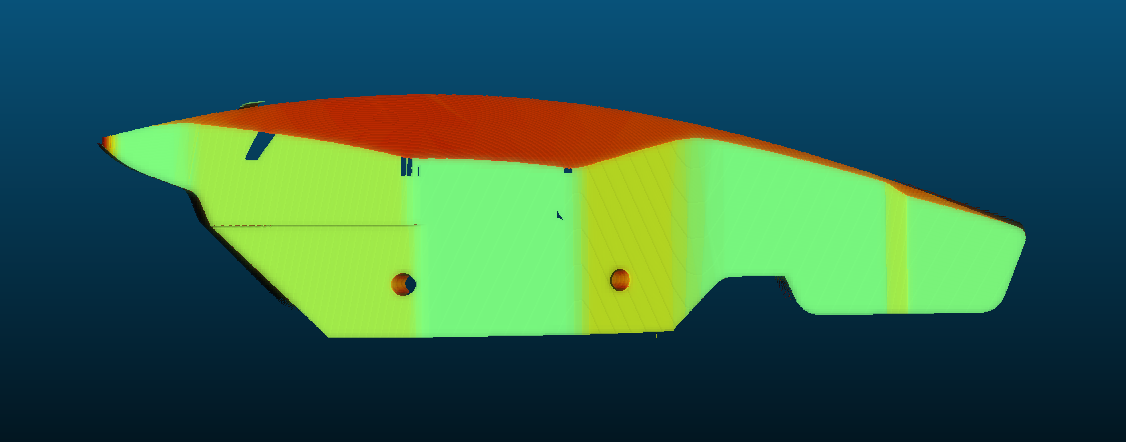}
\includegraphics[width=0.042\columnwidth]{figuras/colorbar.png}}
\caption{Orientation error map showing angular deviations from the optimal sensor orientation at each point, normalized based on defined orientation values. (a) Using trajectory that adapt to the surface of the piece, calculated by the RL algorithm. (b) Using straight trajectory defined by a start point and an end point.}
\label{fig:expDroneSimu2}
\end{figure}


\subsection{Pen Holder}

\subsubsection{Real-world Validation Experiment}

By leveraging the proposed simulation framework to generate and validate the scanning trajectories from the CAD model, the experiment aims to transition these optimized paths effectively to the real-world setup. This process ensures that the developed methods can be reliably applied to the actual drone body, achieving high precision and efficiency in industrial inspection scenarios.

The results obtained after executing the generated trajectory for scanning the Parrot drone in a real environment are presented below. The scanning system is composed of a 6 dof UR3e robotic arm equipped with a triangulation laser profilometer model AT-C5-4090CS30-495. The complete configuration of the inspection system, which includes the UR3e robotic arm, the profilometric sensor, and the drone body, can be seen in Figure 



We begin with a trajectory composed by sensor poses. Using the RoboDK framework [ref]
, we generate the motion program for the UR3e robot to follow the optimized path. This program is then transferred to the robot's controller, configured to execute the motion commands. As a result, the UR3e robotic arm tracks the planned trajectory to scan the Parrot drone, faithfully replicating the precise movements defined during simulation.



\section{Discussion}


In the real-world validation tests of the scanning trajectories, we used a smaller piece that required only a single linear scanning path, eliminating the need for a boustrophedon pattern. This choice was dictated by the physical limitations of the UR3e robotic arm, which has a relatively restricted reach. Given the limited range of the UR3e, it was not feasible to scan larger surfaces that would necessitate multiple passes. Consequently, we opted for a smaller and more suitable piece that allowed for effective validation within the operational capabilities of the available equipment. This decision ensures that, although the UR3e cannot cover extensive surfaces, the methodology and trajectory optimization principles we developed are applicable and verifiable in a controlled and representative environment.

Despite its smaller size, the real piece chosen for the tests has sufficient geometric diversity to validate our proposed methodology. Its varied surface features ensured that the trajectory optimization and scanning techniques could be robustly tested. If we were to work with a larger piece, a boustrophedon scanning pattern would simply be employed. In such cases, our reinforcement learning model would be applied to each pass within the boustrophedon path, just as it was used in the simulated experiment with the door. This approach ensures that our methods are versatile and can be adapted to both small and large-scale scanning tasks.

The scanning trajectory generated by our approach is designed to be highly versatile and adaptable to any robotic system with sufficient reach. This versatility arises from the fact that the trajectory increments in both position and orientation are small and precise enough to be executed by any industrial robot. When these incremental commands are input into the robot’s control software, it calculates the necessary joint movements to achieve the specified end-effector positions.

This adaptability is a key feature of our method. By focusing on end-effector coordinates, our system avoids the need for specific kinematic adjustments for different robots, making the trajectory compatible with a wide range of robotic platforms. Whether the robot has a simple or complex joint configuration, the internal control software translates the end-effector trajectory into joint movements effectively.

This approach not only simplifies the integration process with various robots but also ensures that the trajectory optimization remains effective regardless of the robot used. This makes our system particularly beneficial in diverse industrial settings where different robots may be utilized for various tasks. Thus, our trajectory generation and optimization method provides a flexible solution, generalizing across multiple types of robotic systems and enhancing the applicability of automated inspection processes.


\section{Conclusions}

In this paper, we have presented a method for generating inspection trajectories using laser profilometric sensors using Reinforcement Learning (RL) techniques. Our goal was to optimize the scanning process by dynamically adjusting the sensor's position and tilt to maintain an optimal pose relative to the surface of the part being inspected. 

We employed a simulated environment that replicates the conditions of a real system, as developed in our previous work. In our method, the state space was defined using the position and orientation of the sensor, that normally is the robot's end effector, which allows for the approach to be generalized and easily transferred to different robotic configurations. We also included additional parameters in the state space, such as the mean profile distance, the direction angle, and the spacing between consecutive scans, providing a comprehensive understanding of the inspection process.

The action space was designed to include relative increments in the sensor’s position and tilt angle, allowing for precise adjustments and smooth sensor movements. Our reward function was design by incorporating three key components: the distance between the sensor and the surface, the alignment of the sensor with the surface normal, and the spacing between consecutive scans. This detailed reward function encourages optimal behavior in terms of maintaining appropriate distance, correct orientation, and efficient sensor progression across the surface.

Through our experiments in a simulated environment, we validated the capability of the RL-trained model to adapt to different parts and maintain optimal scanning trajectories not seen during the training phase. Furthermore, we tested the method in a real-world scenario using a UR3e robotic arm. The optimized trajectory, generated offline from a CAD model, was executed successfully, demonstrating that our method can produce high-quality, precise inspection trajectories that ensure effective surface coverage.

\bibliographystyle{IEEEtran}
\bibliography{IEEEabrv,RL_TASE}

\begin{IEEEbiography}[{\includegraphics[width=1in,height=1.25in,clip,keepaspectratio]{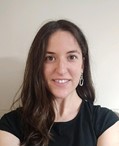}}]{Sara Roos-Hoefgeest}
	Sara Roos Hoefgeest is currently a PhD student at University of Oviedo, in Escuela Politécnica de Ingeniería of Gijón, Spain. She is working at the Department of Electrical, Electronics, Communications, and Systems Engineering. Her research work is focused on the area of Robotics, 3D Computer Vision and Visual Inspection of Manufactured Products. She holds a Bachelor's Degree in Electronic and Automation Engineering and a Master's Degree in Automation and Industrial Computing both from the University of Oviedo.
\end{IEEEbiography}

\begin{IEEEbiography}[{\includegraphics[width=1in,height=1.25in,clip,keepaspectratio]{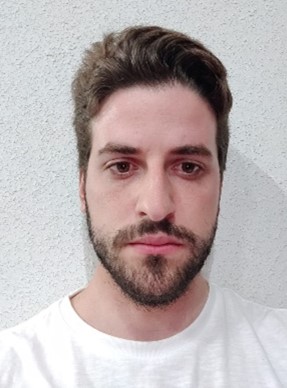}}]{Mario Roos-Hoefgeest}
	Mario Roos Hoefgeest Toribio is currently a PhD student at University of Oviedo, in Escuela Politécnica de Ingeniería of Gijón, Spain. He is working at CIN Systems, applying the results of the research in production lines in industrial facilities. His research work is focused on the area of 3D Computer Vision and Visual Inspection for defects detection in production lines. He holds a Bachelor's Degree in Electronic and Automation Engineering and a Master's Degree in Automation and Industrial Computing both from the University of Oviedo.
\end{IEEEbiography}

\begin{IEEEbiography}[{\includegraphics[width=1in,height=1.25in,clip,keepaspectratio]{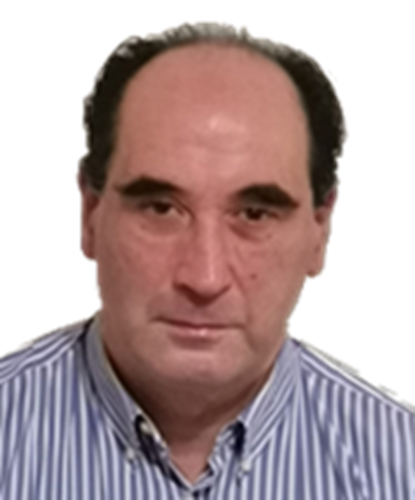}}]{Ignacio Alvarez}
	Ignacio Alvarez is PhD in Industrial Engineering at University of Oviedo (1997), and Associate Professor at the Electrical, Electronics and Automatic Control Department of the same University (1999). His research work is focused in 3D automatic inspection systems for dimensional control and defects detection in production lines, using several technologies (laser triangulation, stereo vision, conoscopic holography). He has participated in more than 20 public financed projects, more than 30 private financed projects, and published more than 15 papers in scientific journals; several prototypes developed or inspired by the author are in production in industrial facilities.
\end{IEEEbiography}

\begin{IEEEbiography}[{\includegraphics[width=1in,height=1.25in,clip,keepaspectratio]{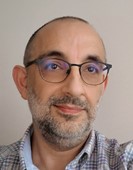}}]{Rafael C. González}
	Rafael Corsino González de los Reyes was born in Gijón, Spain, in 1968. He received an Engineering degree and a Ph.D. degree in Electronics and Automation from the Escuela Superior de Ingenieros Industriales de Gijón, University of Oviedo, Spain, in 1993 and 1999, respectively. He is currently an Associate Professor in the Department of Electrical, Electronics, Communications, and Systems Engineering, Universidad de Oviedo. He is a member of IEEE and his research interests include 3D Computer Vision, Robotic Path Planning and Visual Inspection of Manufactured Products.
\end{IEEEbiography}

\end{document}